\documentclass{article}

\usepackage[preprint]{corl_2021} %

\usepackage{acronym}
\usepackage{graphicx} 
\usepackage{amsmath}
\usepackage{amssymb}
\usepackage{calc} %
\usepackage{bbm}
\usepackage{varwidth}
\usepackage{soul}
\usepackage{layouts}

\usepackage{array} %
\newcolumntype{x}[1]{>{\centering\let\newline\\\arraybackslash\hspace{0pt}}p{#1}}
\usepackage{makecell,booktabs,multirow}

\usepackage{caption}

\usepackage{subcaption}
\newlength{\twosubht}
\newsavebox{\twosubbox}

\usepackage{hyperref}

\usepackage{soul}

\acrodef{2d}[2D]{two dimensional}
\acrodef{3d}[3D]{three dimensional}
\acrodef{6dof}[6DoF]{six degrees of freedom}
\acrodef{ar}[AR]{augmented reality}
\acrodef{cnn}[CNN]{convolutional neural network}
\acrodef{dnn}[DNN]{deep neural network}
\acrodef{icp}[ICP]{iterative closest point}
\acrodef{mvs}[MVS]{multi-view stereo}
\acrodef{rgbd}[RGB-D]{red, green, blue - depth}
\acrodef{sdf}[SDF]{signed distance function}
\acrodef{slam}[SLAM]{simultaneous localization and mapping}
\acrodef{tsdf}[TSDF]{truncated signed distance function}
\acrodef{vo}[VO]{visual odometry}
\acrodef{mvsnet}[CVA-MVSNet]{Cascade View-Aggregation MVSNet}

\acused{rgbd}
\acused{2d}
\acused{3d}
\acused{cnn}
\acused{slam}

\newcommand{\MakeUppercaseSec}[1]{\texorpdfstring{\MakeUppercase{#1}}{#1}}

\newcommand{\vc}[1]{\mathbf{#1}}
\newcommand{\R}{\mathbb{R}}

\newcommand{\pose}{\vc{T}}

\newcommand{\img}{I}
\newcommand{\depth}{\vc{D}}

\newcommand{\feat}{\vc{F}}
\newcommand{\featVol}{\vc{V}}
\newcommand{\depthHyp}{\vc{D}_{hyp}}

\newcommand{\costVol}{\vc{C}}
\newcommand{\probVol}{\vc{P}}
\newcommand{\vaW}{\vc{W}}

\newcommand{\ind}{\mathbbm{1}}

\newlength\aetmplength

\title{TANDEM: Tracking and Dense Mapping\\in Real-time using Deep Multi-view Stereo}

\author{%
Lukas Koestler\textsuperscript{1*}\hspace{8pt}%
Nan Yang\textsuperscript{1,2*,$\dagger$}\hspace{8pt}%
Niclas Zeller\textsuperscript{2,3}\hspace{8pt}%
Daniel Cremers\textsuperscript{1,2}\\[2ex]
\small $^{*}$equal contribution \qquad $^{\dagger}$corresponding author \\[2mm]
\textsuperscript{1}Technical University of Munich\hspace{16pt}\textsuperscript{2}Artisense\\\textsuperscript{3}Karlsruhe University of Applied Sciences
}

\begin{document}
\maketitle

\vspace{-2mm}
\begin{abstract}
In this paper, we present TANDEM, a real-time monocular tracking and dense mapping framework.
For pose estimation, TANDEM performs photometric bundle adjustment based on a sliding window of keyframes.
To increase the robustness, we propose a novel tracking front-end that performs dense direct image alignment using depth maps rendered from a global model that is built incrementally from dense depth predictions.
To predict the dense depth maps, we propose \ac{mvsnet} that utilizes the entire active keyframe window by hierarchically constructing 3D cost volumes with adaptive view aggregation to balance the different stereo baselines between the keyframes.
Finally, the predicted depth maps are fused into a consistent global map represented as a \ac{tsdf} voxel grid.
Our experimental results show that TANDEM outperforms other state-of-the-art traditional and learning-based monocular \ac{vo} methods in terms of camera tracking.
Moreover, TANDEM shows state-of-the-art real-time 3D reconstruction performance.
Webpage: \url{https://go.vision.in.tum.de/tandem}
\end{abstract}


\section{Introduction}\label{sec:introduciton}

Real-time dense \ac{3d} mapping is one of the major challenges in computer vision and robotics.
This problem, known as dense \ac{slam}, includes both estimating the \acs{6dof} pose of a sensor and a dense reconstruction of the surroundings.
While there exist numerous well-working and robust \ac{rgbd} mapping solutions~\cite{Newcombe2011a, Kerl2013, Whelan2015}, real-time dense reconstruction from monocular cameras is a significantly more difficult challenge as depth values cannot be simply read out from the sensor and fused.
Nevertheless, it is a very important problem, as monocular approaches offer significant advantages over \ac{rgbd}-based methods ~\cite{Newcombe2011a} which are usually limited to indoor environment due to the near-range sensing or LiDAR-based~\cite{Zhang2015} solutions which are expensive and heavyweight.

\begin{figure}[t]
  \centering
  \includegraphics[width=\textwidth]{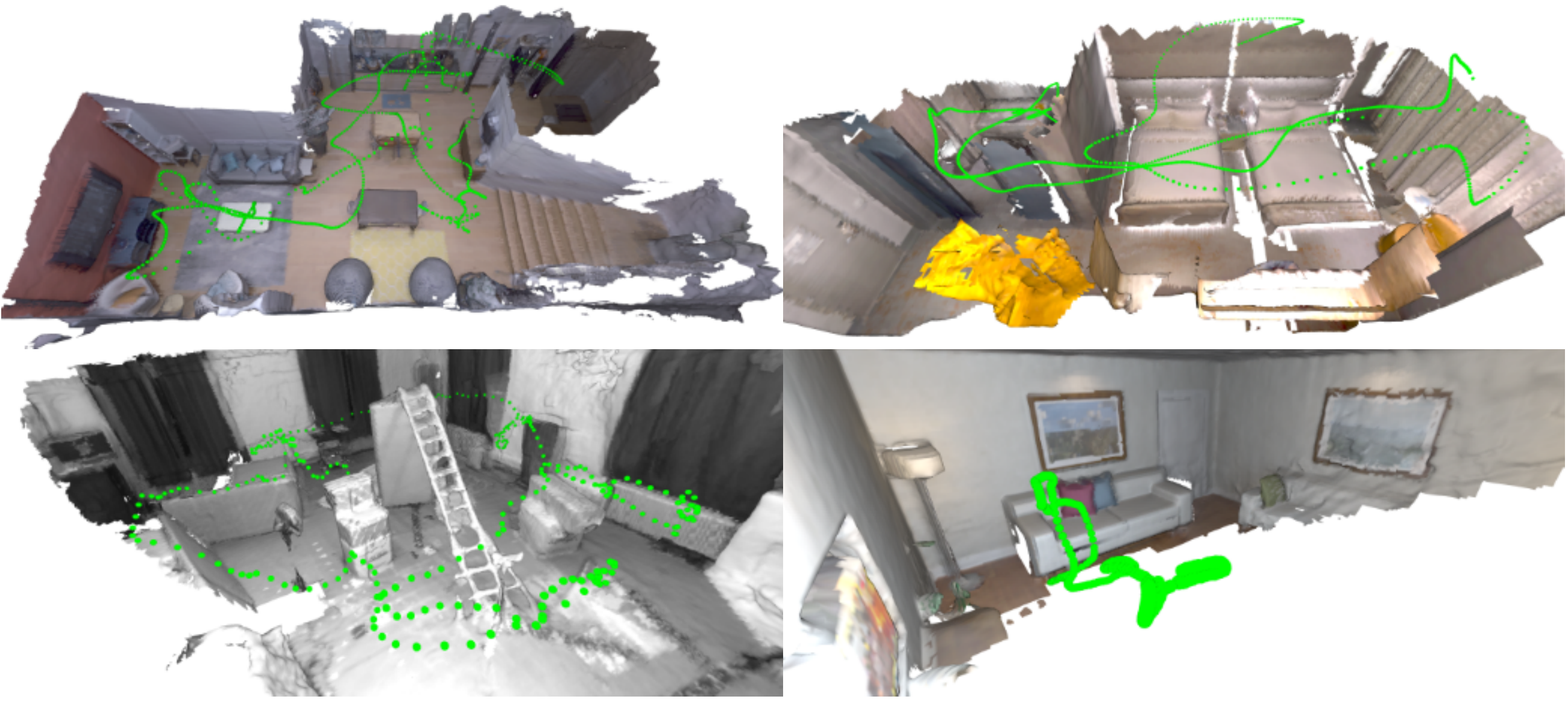}
  \caption{TANDEM is a monocular dense SLAM method that estimates the camera poses and reconstructs the 3D environment in real-time. The figure shows the estimated camera trajectories and the dense 3D models on the sequences of our Replica~\cite{replica19arxiv} test split~(UL, UR), EuRoC~\cite{burri2016euroc}~(BL), and ICL-NUIM~\cite{HandaWMD14}~(BR) using a model trained on the synthetic Replica dataset.}
  \label{fig:teaser}
  \vspace{-0.2cm}
\end{figure}

Several \ac{dnn} based approaches have been proposed to tackle the problem of monocular tracking and dense mapping by utilizing monocular depth estimation~\cite{cnn_slam}, variational auto-encoders~\cite{Bloesch2018code,deep_factors,zuo2020codevio}, or end-to-end neural networks~\cite{murez2020atlas,tang2018ba}.
Unlike the aforementioned works, in this paper, we propose a novel monocular dense \ac{slam} method, TANDEM, which, for the first time, integrates learning-based \ac{mvs} into a traditional optimization-based \ac{vo}. This novel design of dense \ac{slam} shows state-of-the-art tracking and dense reconstruction accuracy as well as strong generalization capability on challenging real-world datasets with the model trained only on synthetic data. \autoref{fig:teaser} shows the 3D reconstructions delivered by TANDEM on unseen sequences.

\textbf{Our contributions.} \textbf{(1)} a novel real-time monocular dense \ac{slam} framework that seamlessly couples classical direct \ac{vo} and learning-based \ac{mvs} reconstruction; \textbf{(2)} to our knowledge, the first monocular dense tracking front-end that utilizes depth rendered from a global \ac{tsdf} model; \textbf{(3)} a novel \ac{mvs} network, \ac{mvsnet}, which is able to leverage the entire keyframe window by utilizing view aggregation and multi-stage depth prediction; \textbf{(4)} state-of-the-art tracking and reconstruction results on both synthetic and real-world data.

\section{Related Work}\label{sec:related_work}

There are two different work streams related to the proposed method. On one side, there is pure \ac{3d} reconstruction based on posed images and, on the other side, there are full \ac{slam} or \ac{vo} frameworks that simultaneously estimate camera poses and a \ac{3d} reconstruction of the environment.

\textbf{3D Reconstruction.}
Most dense \ac{3d} reconstruction approaches consider images and corresponding reference poses as inputs and estimate a dense or partially dense reconstruction of the environment.
Over the last decade, several classical methods have been proposed  
\cite{Furukawa2010, Stuehmer2010, Schoenberger2016}.

Recently, deep-learning-based methods have shown superior performance over classical methods.
These methods regress a \ac{3d} model of the environment utilizing \acp{dnn}.
This \ac{3d} model can be either in the form of a volumetric representation 
\cite{murez2020atlas, weder2020routedfusion, Ji2017surfacenet, Kar2017}, a 
\ac{3d} point cloud \cite{Chen2019} or a set of depth maps 
\cite{huang2018deepmvs, Yao2018mvsnet, zhou2018deeptam, hou2019multi}.
Nowadays, most popular are methods which predict the final model from \ac{3d} cost volumes.
Huang et al.~\cite{huang2018deepmvs} proposed one of the first cost-volume-based approaches.
Zhou et al.~\cite{zhou2018deeptam} aggregate multiple image-pair-wise volumes to a single cost volume and use a \acs{3d} \acs{cnn} for depth prediction.
Yao et al.~\cite{Yao2018mvsnet} propose to directly calculate a single volume based on \ac{2d} deep feature maps predicted from the input images.
In a follow-up work, Yao et al.~\cite{Yao2019recmvsnet} replace the depth prediction \acs{cnn} by a recurrent network.
To improve run-time and memory consumption, Gu et al.~\cite{Gu2020cascade} propose a cascade cost volume.
Yi et al.~\cite{yi2019PVAMVSNET} introduce a self-adaptive view aggregation to weigh the voxel-wise contribution for each input image. The proposed \ac{mvsnet} is built upon the two aforementioned works~\cite{Gu2020cascade,yi2019PVAMVSNET} and largely inspired by them. However, only by their combination and adaption to the \ac{slam} setting we achieve better performance and real-time capability.

While all previous methods are based on a set of selected frames, Murez et al.~\cite{murez2020atlas} instead directly predict a \acs{tsdf} model from a single global \ac{3d} cost volume.
Weder et al.~\cite{weder2020routedfusion} propose a learning-based alternative to classical \acs{tsdf} fusion of depth maps. 
While these volumetric representations, in general, are rather memory intense, Niessner et al.~\cite{Niesner2013} propose voxel-hashing to overcome this limitation and Steinbr{\"{u}}cker et al.~\cite{DBLP:conf/icra/SteinbruckerSC14} perform depth map fusion on a CPU using an octree.

\textbf{\ac{rgbd} SLAM.}
In the area of visual \ac{slam}, \ac{rgbd} approaches by nature provide dense depth maps along with the camera trajectory and therefore target to solve a similar problem as our approach.
Bylow et al.~\cite{Bylow2013} and Kerl et al.~\cite{Kerl2013,Kerl2013a} mainly focus on accurate trajectory estimation from \ac{rgbd} images. In addition to camera tracking, Newcombe et al.~\cite{Newcombe2011a} integrate the depth maps into a global \ac{tsdf} representation.
Whelan et al.~\cite{Whelan2015} perform surfel-based fusion and non-rigid surface deformation for globally consistent reconstruction.
K{\"a}hler et al.~\cite{Kaehler2016} use a \ac{tsdf} map representation which is split into sub maps to model loop closures.
While most previous methods optimize only for the frame pose, Sch{\"o}ps et al.~\cite{Schoeps2019} propose a full bundle adjustment based direct \ac{rgbd} \ac{slam} which optimizes for both camera pose and \ac{3d} structure.
Sucar et al. \cite{Sucar:etal:iccv2021} integrate a \ac{dnn}-based implicit scene representation into an \ac{rgbd} \ac{slam} system. 

\begin{figure}[t]
\sbox\twosubbox{%
  \resizebox{\dimexpr.99\textwidth-1em}{!}{%
    \includegraphics[height=3cm]{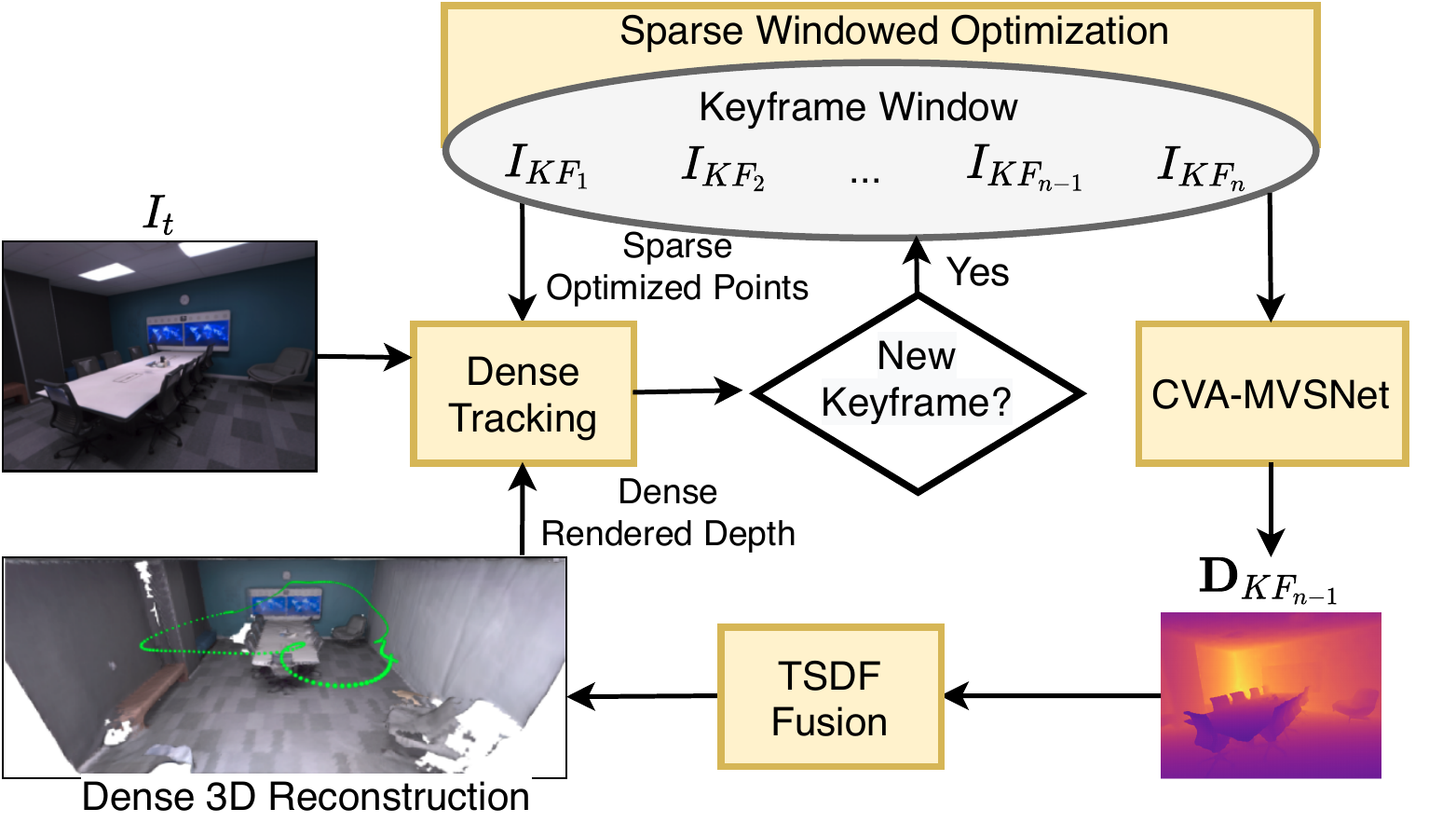}%
    \includegraphics[height=3cm]{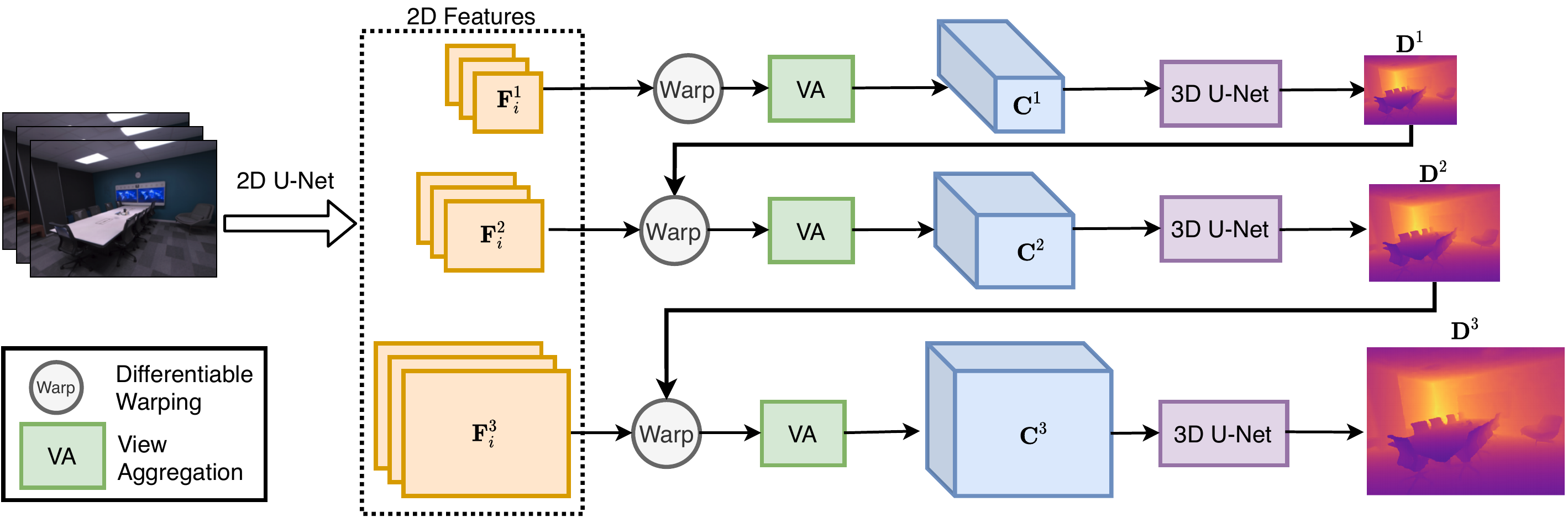}%
  }%
}
\setlength{\twosubht}{\ht\twosubbox}

\centering
\subcaptionbox{TANDEM System Overview.\label{fig:system_a}}{%
  \includegraphics[height=\twosubht]{figures/corl_system_overview.pdf}%
}\quad
\subcaptionbox{\ac{mvsnet} Architecture.\label{fig:system_b}}{%
  \includegraphics[height=\twosubht]{figures/corl_tandem_mvs.pdf}%
}
\caption{(a) Every new frame is tracked using the optimized sparse points from the visual odometry and the dense depth map rendered from the 3D model. The poses of the keyframes are estimated by sliding-window photometric bundle adjustment and fed into the \ac{mvsnet} for dense depth prediction. The depth maps are fused into a globally consistent \ac{tsdf} volume. (b) \ac{mvsnet} builds cascaded cost volumes and hierarchically estimates the depth maps. The view aggregation module effectively aggregates the features of the multi-view images by predicting self-adaptive weights.
}
\label{fig:system}
\vspace{-0.2cm}
\end{figure}

\textbf{Monocular SLAM.}
Compared to \ac{rgbd} methods, for monocular approaches both tracking and mapping become much more challenging.
Using a single monocular camera, Newcombe et al.~\cite{Newcombe2011dtam} 
perform optimization based on a photometric cost volume to jointly estimate the 
camera pose and dense depth maps in real-time on a GPU. Pizzoli et 
al.~\cite{Pizzoli2014remode} combine a depth Bayesian 
filter~\cite{vogiatzis2011video} with the spatial smoothness prior.
Engel et al. \cite{lsd_slam} propose the first large-scale photometric \ac{slam} formulation including loop closure detection and pose graph optimization.
By using a sparse representation, Engel et al.~\cite{Engel2018dso} were able to formulate the first fully photometric \ac{vo} framework which jointly estimates pose and depth in real-time.
To obtain denser reconstructions, Mur-Artal et al.~\cite{Mur-Artal2015} perform semi-dense probabilistic depth estimation on top of feature-based \ac{slam}~\cite{MurArtal2015}.
Sch{\"o}ps et al.~\cite{Schoeps2017a} perform temporal, plane-sweep-based depth estimation using the poses and images obtained from a mobile tracking device.
Tateno et al.~\cite{cnn_slam} and Yang et al.~\cite{yang2018deep,Yang2020d3vo} leverage \acp{dnn} in a traditional direct \ac{slam} framework to improve tracking performance and overcome the problem of scale ambiguity. 
While traditional geometric and hybrid approaches still achieve superior tracking performance, there are several fully learned \ac{slam} frameworks~\cite{Zhou2017, Yin2018geonet, Zhan2018, zhou2018deeptam}, which are superior in terms of reconstruction completeness.
Jatavallabhula et al.~\cite{jatavallabhula2020gradslam} propose a differential optimizer which has the potential to bridge the gap between traditional and learning-based \ac{slam}.
A novel idea for deep-learning-based \ac{slam} is proposed by Bloesch et al.~\cite{Bloesch2018code}.
The authors propose to learn a frame-wise code representation for the scene depth, which is the jointly optimized together with the camera pose.
The work of Czarnowski et al.~\cite{deep_factors} is an extension of \cite{Bloesch2018code}, where the code representation is integrated into a full \ac{slam} system.
Zuo et al. \cite{zuo2020codevio} make use of a similar code representation in a visual-inertial setup and furthermore feed sparse, tracked features into a \ac{dnn}.

\section{\MakeUppercaseSec{TANDEM}}\label{sec:system}
The proposed TANDEM is comprised of three components: monocular visual odometry (\autoref{sec:visual_odometry}), dense depth estimation with \ac{mvsnet} (\autoref{sec:depth_estimation}), and volumetric mapping (suppl.). \autoref{fig:system_a} shows an overview of the system. The visual odometry utilizes the monocular video stream and the dense depth rendered from the 3D \ac{tsdf} model to estimate camera poses in a sliding-window manner. Given the keyframes and their estimated poses, the proposed \ac{mvsnet} predicts a dense depth map for the reference keyframe. To reconstruct a complete and globally consistent 3D model of the environment, the depth maps are then fused into the \ac{tsdf} voxel grid~\cite{CurlessLevoy96} with voxel-hashing~\cite{Niesner2013}. By seamlessly integrating these components, the resultant system TANDEM enables real-time tracking and high-quality dense mapping from a monocular camera. Further details, including on the \ac{tsdf} volume initialization, are given in the supplementary material.

\subsection{Visual Odometry}\label{sec:visual_odometry}
Estimating camera poses by tracking a sparse set of points across multiple frames has shown great performance in recent \ac{vo} systems~\cite{Engel2018dso,MurArtal2015}.
Using more points for the joint optimization, however, does not necessarily further improve the accuracy of the estimated poses while significantly increases the runtime~\cite{Engel2018dso}.
Therefore, in the proposed \ac{vo} system we make use of a direct sparse windowed optimization back-end as described in Direct Sparse Odometry (DSO)~\cite{Engel2018dso}.
However, we utilize dense depth maps rendered from the global \ac{tsdf} model, which we build up incrementally, in the direct image alignment front-end.
In numerous experiments, we confirm that this combination of dense tracking front-end and sparse back-end optimization significantly improves the tracking performance (cf. \autoref{tab:trajectory_eval}) while maintaining a fast runtime.

\textbf{Dense Front-end Tracking.}
The front-end tracking provides camera-rate pose estimations and serves as initialization for the windowed optimization back-end.
In the original DSO, a new frame is tracked against the last keyframe $n$ by direct image alignment using a sparse depth map $\depth_{n}^\text{DSO}[\vc{p}]$ generated from all points in the optimization window.
This approach, however, lacks robustness (cf. \autoref{tab:trajectory_eval}) due to the sparsity of the depth map.
We alleviate the issue by incorporating a dense depth map $\depth_{n}^\text{TSDF}$ which is rendered from the constructed \ac{tsdf} model.
For each pixel $\vc{p}$ in the current keyframe $n$, we assign a depth value either based on the sparse \ac{vo} points $\depth_{n}^\text{DSO}[\vc{p}]$, if available, or based on the rendered dense depth $\depth_{n}^\text{TSDF}[\vc{p}]$, otherwise.
Due to the incrementally-built \ac{tsdf} model, the combined depth buffer might not contain valid depth values for all pixels but it is much denser in comparison to using the sparse depth values only.
The nearly-dense combined depth map is used for two-frame direct image alignment.

\subsection{CVA-MVSNet}\label{sec:depth_estimation}

Let $\{(\img_i, \pose_i)\}_{i=1}^n$ be the set of active keyframes where $\img_i$ is the image of size $(H, W)$ and $\pose_i$ is the corresponding global pose estimated by the \ac{vo}. \ac{mvsnet} is based on the principles of \acl{mvs}~\cite{furukawa2015multi} and further leverages deep neural networks~\cite{Yao2018mvsnet} to estimate a depth map for the reference frame $\img_{n-1}$. \ac{mvsnet} overcomes the prohibitive memory requirement of deep \ac{mvs} networks by hierarchically estimating the depth using cascaded cost volumes and aggregates the deep features of all the keyframes effectively with a self-adaptive view aggregation module. 

As shown in \autoref{fig:system_b}, the multi-scale deep features $\feat_i^s$ of the keyframes are firstly extracted by 2D U-Nets with shared weights, where $i\in [1,n]$ is the frame index and $s \in [1, 3]$ is the scale index. As a result, $\feat_i^s$ is of the shape $(F^s, H^s, W^s)$ where $F^s$ is the feature dimension of the scale $s$, $H^s=H/2^{3-s}$, and $W^s=W/2^{3-s}$. The depth map of the reference frame is estimated hierarchically with 3 stages each of which takes the set of features $\{\feat_i^s\}_{i=1}^{n}$ as the inputs and predicts the reference depth map $\depth^s$ of shape $(H^s, W^s)$. For clarity, we first explain how a single stage estimates the depth and then describe how multiple stages are assembled hierarchically.

\textbf{Single Stage Depth Estimation.}
For each stage, a cost volume $\costVol^s$ needs to be constructed using the deep features $\{\feat_i^s\}_{i=1}^{n}$. For each pixel of the reference frame, we define $D^s$ depth hypotheses, which results in a tensor $\depthHyp^s$ of shape $(D^s, H^s, W^s)$. The deep features $\feat_i^s$ of each frame are geometrically transformed with differentiable warping~\cite{jaderberg2015spatial} using the depth hypotheses, the relative pose $\pose_j^i = \pose_i^{-1}\pose_j$ and the camera intrinsics. As a result, a feature volume $\featVol_i^s$ of shape $(F^s, D^s, H^s, W^s)$ is constructed for every frame.

In order to aggregate the information from multi-view feature volumes into one cost volume $\costVol^s$, most prior deep MVS methods treat different views equally and use a variance-based cost metric:
\begin{equation}
    \costVol^s = \frac{\sum_{i=1}^n \, (\featVol_i^s - \bar{\featVol^s})^2}{n} \,, \quad \text{where} \qquad \bar{\featVol} = \frac{\sum_{i=1}^n\featVol_i^s}{n} \, .
\end{equation}
However, in the sliding-window \ac{slam} setting, the keyframes are not evenly distributed within the optimization window -- typically the distance between newer keyframes is much smaller than between older keyframes. This causes considerable occlusion and non-overlapping images. The variance-based cost volume, which weighs different views equally, is thus inappropriate. To alleviate this issue, we employ self-adaptive view aggregation~\cite{yi2019PVAMVSNET} to construct the cost volume:
\begin{equation}
    \costVol^s = \frac{\sum_{i = 1, i \neq j}^n (1 + \vaW_i^s) \, \odot \, (\featVol_i^s - \featVol_j^s)^2}{n - 1} \, ,
\end{equation}
where the view aggregation weights $\vaW_i^s$ have the shape $(1, D^s, H^s, W^s)$ and $\odot$ is element-wise multiplication with broadcasting. The view aggregation weights $\vaW_i^s$ are estimated by a shallow 3D convolutional network for each $\featVol_i^s$ separately by taking $(\featVol_i^s - \featVol_j^s)^2$ as the input. This aggregation module allows the network to adaptively downweight erroneous information.

The cost volume $\costVol^s$ is then regularized using a 3D U-Net and finally passed through a softmax non-linearity to obtain a probability volume $\probVol^s$ of shape $(D^s, H^s, W^s)$. Given the per-pixel depth hypotheses $\depthHyp^s$ of shape $(D^s, H^s, W^s)$ the estimated depth is given as the expected value
\begin{equation}
    \depth^s[h, w] = \sum_{d = 1}^D \; \probVol^s[d, h, w] \cdot \depthHyp^s[d, h, w].
\end{equation}

\textbf{Hierarchical Depth Estimation.}
The network leverages the depth estimated from the previous stage $\depth^{s-1} (s > 1)$ to define a fine-grained depth hypothesis tensor $\depthHyp^s$ with a small $D^s$. Since no prior stage exists for the first stage, each pixel of $\depthHyp^1$ has the same depth range defined by $[d_{min}, d_{max}]$ with $D^1 = 48$ depth values. For the later stages ($s>1$), the depth $\depth^{s-1}$ is upsampled and then used as a prior to define $\depthHyp^s$. Specifically, for the pixel location $(h, w)$, $\depthHyp^s(\cdot, h, w)$ is defined using the upsampled $\depth^{s-1}(h, w)$ as the center and then sampled $D^s$ values around it using a pre-defined offset~\cite{Gu2020cascade}. In this way, fewer depth planes are needed for the stage with a higher resolution, i.e., $D^1 \geq D^2 \geq D^3$.
We train the network using the $L1$ loss applied on all three stages with respect to the ground-truth depth and use the sum as the final loss function.

\subsection{Implementation Details}
To guarantee real-time execution, TANDEM leverages parallelism on multiple levels. Dense tracking and bundle adjustment are executed in parallel threads on the CPU while, asynchronously and in parallel, TSDF fusion and \ac{dnn} inference are run on the GPU. We train our \ac{mvsnet} in PyTorch~\cite{NEURIPS2019_9015} and perform inference in C++ using TorchScript.

TANDEM can processes images at ca. $20$ FPS while running on a desktop with an Nvidia RTX 2080 super with $8$ GB VRAM, and an Intel i7-9700K CPU. This includes tracking and dense TSDF mapping but no visualization or mesh generation.
We refer to the supplementary for further details, including how to potentially scale the network for deployment on embedded platforms.

\section{Experimental Results}\label{sec:experiments}
As TANDEM is a complete dense SLAM system, we evaluate it for both monocular camera tracking and dense 3D reconstruction. Specifically, for the camera tracking, we compare with state-of-the-art traditional sparse monocular odometry, DSO~\cite{Engel2018dso} and ORB-SLAM~\cite{MurArtal2015}, as well as learning-based dense SLAM methods, DeepFactors~\cite{deep_factors} and CodeVIO~\cite{zuo2020codevio}. For the 3D reconstruction, we compare with a state-of-the-art deep \acl{mvs} method, Cas-MVSNet~\cite{Gu2020cascade}, end-to-end reconstruction method, Atlas~\cite{murez2020atlas}, learning-based dense SLAM methods, CNN-SLAM~\cite{cnn_slam}, DeepFactors~\cite{deep_factors}, and CodeVIO~\cite{zuo2020codevio}, as well as iMAP~\cite{Sucar:etal:iccv2021}, a recently proposed RGB-D dense SLAM method using a deep implicit map representation~\cite{park2019deepsdf,mescheder2019occupancy}.

In the following, we will first introduce the datasets we use for training and evaluation. Note that only \ac{mvsnet} needs to be trained while the dense tracking part of TANDEM is purely optimization-based and does not require training on specific datasets. Then, the ablation study for TANDEM demonstrates different design choices. In the end, we show quantitative and qualitative comparisons with other state-of-the-art methods. Due to the limited space, we demonstrate only part of the conducted evaluation and refer to the supplementary material for additional experiments.

\subsection{Datasets}
\begin{figure*}[t]
	\centering
	\includegraphics[width=1\linewidth]{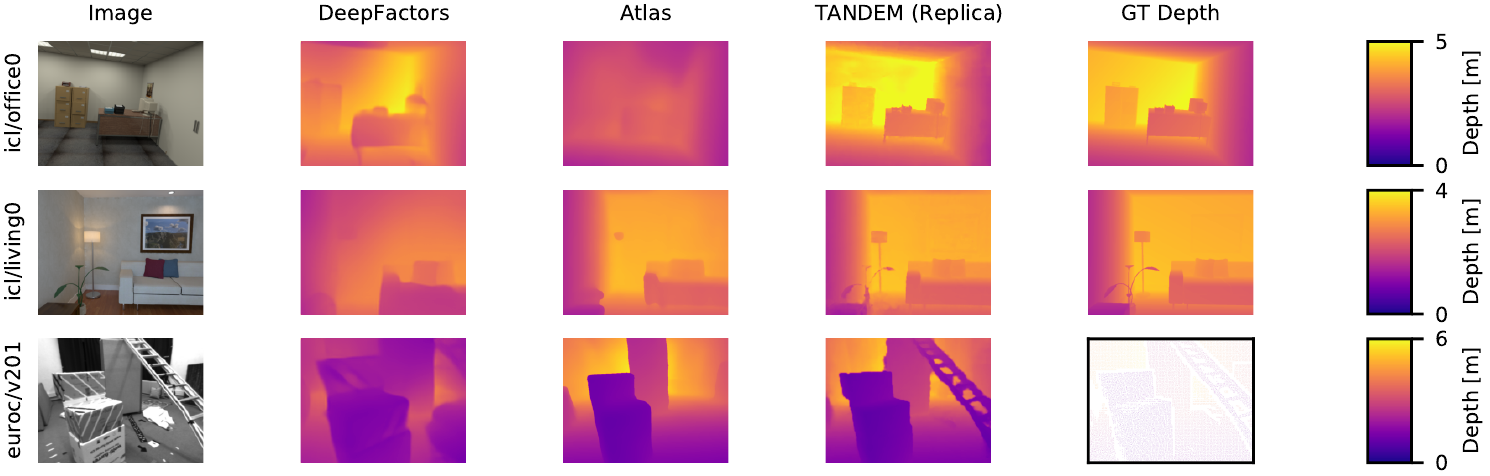}
    \caption{Depth comparison for DeepFactors~\cite{deep_factors}, Atlas~\cite{murez2020atlas}, and TANDEM on unseen sequences. TANDEM produces finer-scale details, e.g. the plant in the second row, or the ladder in the third row.
    For EuRoC only sparse ground-truth depth is available. A high-resolution version of this figure can be found in the supplementary material.
    }
    \label{fig:depth_qual}
    \vspace{-0.2cm}
\end{figure*}
\begin{figure*}[tb]
  \centering
  \includegraphics[width=1\linewidth]{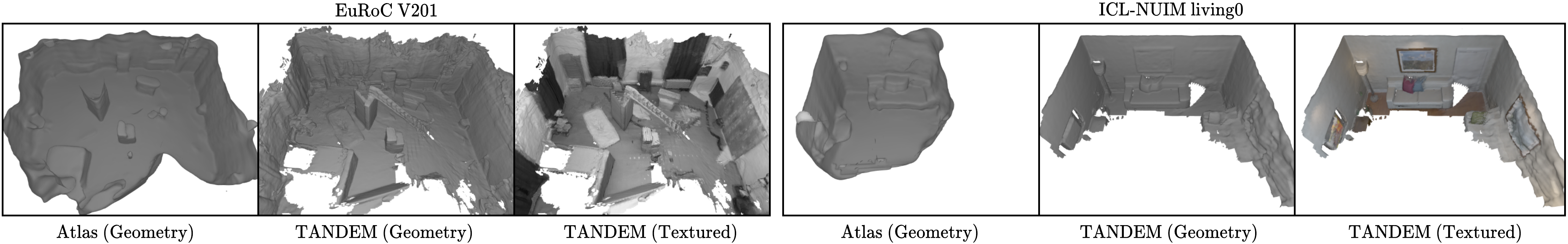}
  \vspace{-0.3cm}
  \caption{Qualitative comparison of Atlas~\cite{murez2020atlas} and TANDEM on unseen sequences. Atlas does not construct textured meshes, so we also render the pure geometry from TANDEM for comparison.
  }
  \label{fig:meshes_qual}
    \vspace{-0.2cm}
\end{figure*}

\textbf{Training sets}. We train two models for \ac{mvsnet}: One on the real-world \textbf{ScanNet}~\cite{dai2017scannet} dataset and the other one on the synthetic \textbf{Replica} dataset~\cite{replica19arxiv}.
The ScanNet dataset consists of $1513$ indoor scenes and we use the official train and test split for training \ac{mvsnet} to give a fair comparisons with DeepFactors and Atlas.
However, the geometry and texture in ScanNet show noticeable artifacts and incompleteness~\cite{replica19arxiv}, which limits its potential to training high-quality dense reconstruction methods. To complement the ScanNet dataset, we build upon the recently proposed Replica dataset~\cite{replica19arxiv}, which consists of $18$ photorealistic scenes. These scenes were captured from real-world rooms using a state-of-the-art 3D reconstruction pipeline and show very high-quality texture and geometry. Because the authors did not release any sequences, we extend the dataset by manually creating realistic camera trajectories that yield 55 thousand poses and images.

\textbf{Evaluation sets}.
We use the \textbf{ICL-NUIM}~\cite{HandaWMD14} dataset and the Vicon Room sequences of the \textbf{EuRoC} dataset~\cite{burri2016euroc} for evaluating the tracking and the dense 3D reconstruction. Note that TANDEM is not trained on either of the datasets. ICL-NUIM is a synthetic indoor dataset with pose and dense depth ground-truth. It contains low-texture scenes which are challenging for monocular visual odometry and dense depth estimation. EuRoC is a real-world dataset recorded by a micro aerial vehicle (MAV).

\subsection{Camera Pose Estimation}
\label{sec:pose_exp}
We evaluate TANDEM for pose estimation against other state-of-the-art monocular SLAM methods on EuRoC and ICL-NUIM. On EuRoC, we evaluate against DSO~\cite{Engel2018dso}, ORB-SLAM2~\cite{MurArtal2015}, DeepFactors~\cite{deep_factors}, and CodeVIO~\cite{zuo2020codevio} uses a camera and an IMU sensor. Note that we turn off the global optimization and relocalization of ORB-SLAM2 for a fair comparison. We also implement a variant of DSO (DSO + Dense Depth) which uses all the pixels of the dense depth maps estimated by \ac{mvsnet} for the front-end direct image alignment. Note that the difference between TANDEM and DSO + Dense Depth is that TANDEM tracks against the \textit{global 3D model} by rendering the depth maps from the TSDF grid. TANDEM achieves overall better tracking accuracy and robustness than the other monocular methods on both ICL-NUIM and EurRoC. Due to the limited space, we show the results on EuRoC in~\autoref{tab:trajectory_eval} of the main paper and kindly refer to the supplementary material for the results on ICL-NUIM where we also show the comparision with DeepTAM~\cite{zhou2018deeptam}.

All the methods except for CodeVIO~\cite{zuo2020codevio} are run five times for each sequence and reported with the mean RMSE error and standard deviation in terms of absolute pose estimations after Sim(3) alignment with ground-truth. For CodeVIO, we directly take the numbers reported in their paper. The comparison with DSO and DSO + Dense Depth indicates that the proposed dense tracking against the global 3D model improves the camera pose estimations, especially on the more challenging sequences (V102 and V202). However, we should admit that TANDEM still cannot compete with CodeVIO which uses an IMU sensor for pose estimations.

\begin{table}[t]
\scriptsize
\caption{Pose evaluation on EuRoC~\cite{burri2016euroc}. All the methods are $\text{Sim} (3)$ aligned w.r.t. the ground-truth trajectories. The mean absolute pose errors and the standard deviations over five runs are shown.}
\label{tab:trajectory_eval}
\begin{center}
\begin{tabular}{p{1.5cm} c | c c c c c }
    \toprule
    Sequence       &
    \makecell{CodeVIO\cite{zuo2020codevio}}  &
    \makecell{DeepFactors\cite{deep_factors}}  &
    \makecell{DSO~\cite{Engel2018dso}} &
    \makecell{ORB-SLAM2~\cite{MurArtal2015}} &
    \makecell{DSO+Dense Depth} &
     Ours  \\
    \midrule
EuRoC/V101    &          \textbf{0.06} (-) &     1.48 ($\pm$0.134)  &             0.10 ($\pm$0.006) &     0.31 ($\pm$0.22\phantom{0})  &   \underline{0.09} ($\pm$0.002)   &      \underline{0.09} ($\pm$0.001)     \\
EuRoC/V102    &          \textbf{0.07} (-) &     Lost                &       0.27 ($\pm$0.017) &     \underline{0.11} ($\pm$0.05\phantom{0})   &     0.28 ($\pm$0.015)   &   0.17 ($\pm$0.006) \\
EuRoC/V201    &          \textbf{0.10} (-) &   1.06 ($\pm$0.441)   &        \underline{0.09} ($\pm$0.005) &    1.40 ($\pm$0.211)   &      \underline{0.09} ($\pm$0.003)   &   \underline{0.09} ($\pm$0.002)      \\
EuRoC/V202    &          \textbf{0.06} (-) & 1.89 ($\pm$0.019)   &       0.21 ($\pm$0.020) &    0.84 ($\pm$0.648) &      0.19 ($\pm$0.022)   &   \underline{0.12} ($\pm$0.009) \\

\bottomrule
\end{tabular}
\end{center}
\vspace{-0.5cm}
\end{table}

\subsection{Ablation Study}
We conduct the ablation study of \ac{mvsnet} on the test split of Replica and show the results in ~\autoref{tab:ablation}. Specifically, we evaluate the effectiveness of using the full VO window with 7 keyframes (Win), the view aggregation module (VA), and fewer depth planes (S) with $(48, 4, 4)$ for $(D^1, D^2, D^3)$. The baseline method is the original Cas-MVSNet using 3 multi-view images as the inputs, no view aggregation module, and more depth planes $(48, 32, 8)$. We use the absolute difference (Abs), and the percentage of inliers with different thresholds ($a_1$, $a_2$, $a_3$) as the metrics for the depth map evaluation. Please refer to the supplementary material for the formulas of the metrics. In addition, we also show the inference time and the memory usage of different models. From the table, we can see that using the entire keyframe window with more frames does not improve the accuracy over the baseline model while increasing the runtime and memory usage. With the view aggregation module, the accuracy is significantly improved, but the runtime further increases. Using fewer depth planes does not show a significant drop in accuracy but improves the runtime and memory usage a lot. Therefore, to guarantee the real-time performance of TANDEM, we use the fewer-plane model as the final \ac{mvsnet} and all other experiments in the paper are conducted with this model.

\begin{table}[tb]
\small
\caption{Ablation study of \ac{mvsnet} on Replica~\cite{replica19arxiv}. Using all keyframes within the \ac{vo} window (\textit{Win}) does not improve the baseline. However, combining \textit{Win} with view aggregation (\textit{VA}) yields more accurate results at the cost of increased inference runtime and memory. By reducing the number of depth planes (\textit{S}) from ${(48, 32, 8)}$ to ${(48, 4, 4)}$ we retain high quality and guarantee the real-time performance of TANDEM. Best shown in \textbf{bold} and second best shown \underline{underlined}.}
\label{tab:ablation}
\begin{center}
\begin{tabular}{p{3.2cm} c c c c c c c c c}
    \toprule
&
Win &
VA &
\makecell{S} &
\makecell{Abs$\downarrow$ \\\relax[cm]} &
\makecell{$a_1\uparrow$\\\relax[\%]} &
\makecell{$a_2\uparrow$\\\relax[\%]} &
\makecell{$a_3\uparrow$\\\relax[\%]} &
\makecell{Time$\downarrow$\\\relax[ms]} &
\makecell{Mem.$\downarrow$\\\relax[MiB]} \\
\midrule
baseline~\cite{Gu2020cascade}                &                      &                       &                       & 2.64                  & \underline{98.64}     & 82.33                 & 20.12                 & \textbf{142}          & \underline{3447}      \\
+ \ac{vo} window          &     \checkmark       &                       &                       & 2.64                  & 98.39                 & 83.55                 & 20.56                 & 215                   & 4117                  \\
+ View aggregation     &     \checkmark       &      \checkmark       &                       & \textbf{1.92}         & \textbf{99.00}        & \textbf{88.59}        & \underline{26.03}     & 288                   & 4117                  \\
+ Fewer depth planes         &     \checkmark       &      \checkmark       &      \checkmark       & \underline{2.33}      & 98.51                 & \underline{86.93}     & \textbf{26.09}        & \underline{158}       & \textbf{2917}         \\
\bottomrule
\end{tabular}
\end{center}
\vspace{-0.5cm}
\end{table}

\subsection{3D Reconstruction}
We evaluate the reconstruction accuracy on both ICL-NUIM and EuRoC. On ICL-NUIM we compare with DeepFactors~\cite{deep_factors}, CNN-SLAM~\cite{cnn_slam}, Atlas~\cite{murez2020atlas}, and Cas-MVSNet~\cite{Gu2020cascade}. \autoref{tab:reconst_eval_icl} shows the evaluation results.
Since Atlas, a pure \ac{3d} reconstruction method, does not estimate poses, we provide ground-truth poses as the input.
Note that Atlas estimates a TSDF volume directly from a 3D \ac{cnn}, so we render the depth maps for evaluation against other methods. CNN-SLAM, DeepFactors, and Atlas are trained on ScanNet. For Cas-MVSNet, we re-train it on the Replica dataset and use the same poses as for our \ac{mvsnet}. We show the results of the evaluations for our ScanNet-trained model and our Replica-trained model to facilitate a fair comparison.
The depth maps of monocular methods are aligned in scale based on the trajectory and further details are given in the supplementary.
We use the $a_1$ metric as the major measurement for accuracy. From \autoref{tab:reconst_eval_icl}, we can see that our method shows a notable improvement in comparison to the other methods and delivers the best result on average. Note that \ac{mvsnet} achieves better results with the model trained on the synthetic Replica than the model trained on ScanNet.

\begin{table}[tb]
\small
\caption{Depth evaluation on ICL-NUIM~\cite{HandaWMD14}. We show the percentage of pixels for which the estimated depth falls within 10\% of the groundtruth value.
}
\label{tab:reconst_eval_icl}
\begin{center}
\begin{tabular}{p{1.2cm} c c c c c c}
    \toprule
    Sequence       &
    \makecell{CNN-SLAM\\\cite{cnn_slam}} &
    \makecell{DeepFactors\\\cite{deep_factors}} &
    \makecell{Atlas\\\cite{murez2020atlas}} &
    \makecell{Cas-MVSNet\\\cite{Gu2020cascade}} &
    \makecell{Ours\\(ScanNet)} &
    \makecell{Ours\\(Replica)} \\
    
    \midrule
    icl/office0    & 19.41 & 30.17             & 28.79            & \underline{77.73} & 52.34             & \textbf{84.04}      \\
    icl/office1    & 29.15 & 20.16             & 62.89            & \underline{88.87} & 61.83             & \textbf{91.18}      \\
    icl/living0    & 12.84 & 20.44             & 83.16            & \textbf{97.06} & 86.42             & \underline{97.00}      \\
    icl/living1    & 13.04 & 20.86             & 36.93            & \underline{84.11} & 71.35             & \textbf{90.62}      \\
    \midrule
    \textbf{Average}  & 19.77 & 30.17             & 66.93            & \underline{86.94} & 67.99            & \textbf{90.71}      \\
    \bottomrule
\end{tabular}
\end{center}
\vspace{-0.5cm}
\end{table}
\begin{table}[tb]
\vspace{-0.2cm}
\small
\caption{Depth evaluation on EuRoC~\cite{burri2016euroc}. We show the percentage of correct pixels $d_1$ as in \cite{eigen_split}.
}
\label{tab:reconst_eval_euroc}
\begin{center}
\begin{tabular}{p{1.7cm} c c c c c c}
    \toprule
    Sequence       &
    \makecell{CodeVIO\cite{zuo2020codevio}} &
    \makecell{DeepFactors\cite{deep_factors}} &
    \makecell{Atlas\cite{murez2020atlas}} &
    \makecell{Cas-MVSNet\cite{Gu2020cascade}} &
    \makecell{Ours\\(ScanNet)} &
    \makecell{Ours\\(Replica)} \\

    \midrule
    EuRoC/V101    & 86.99     & 71.82             & 57.16            & 93.05         & \underline{93.69}    & \textbf{94.25}      \\
    EuRoC/V102   & 78.65      & X             & \textbf{92.29}      & 88.03     & 89.62             & \underline{90.50}      \\
    EuRoC/V201    & 77.32     & 71.85             & 93.28            & 95.43         & \underline{96.84}   & \textbf{97.17}      \\
    EuRoC/V202    & 71.98     & 68.26             & 64.33            & \underline{93.16}         & 92.65   & \textbf{95.68}      \\
    \midrule
    \textbf{Average}  & 78.74 &              & 76.77            & 92.42         & \underline{93.20}            & \textbf{94.40}      \\
    \bottomrule
\end{tabular}
\end{center}
\vspace{-0.5cm}
\end{table}

\begin{table}[h!]
\small
\caption{Comparison to iMAP~\cite{Sucar:etal:iccv2021}. TANDEM shows comparable performance to iMAP, which uses RGB-D data, but no training prior to scanning.
The mesh-based 3D metrics are as in \cite{Sucar:etal:iccv2021}.}
\label{tab:imap}
\begin{center}
\begin{tabular}{p{2.4cm} c c c c @{\hspace{0.2cm}} c c c}
\toprule
&
\multicolumn{3}{c}{room-1} & &
\multicolumn{3}{c}{office-4} \\

\cmidrule{2-4}
\cmidrule{6-8}

&
\makecell{Acc [cm]}&
\makecell{Comp [cm]}&
\makecell{CR [\%]}& &
\makecell{Acc [cm]}&
\makecell{Comp [cm]}&
\makecell{CR [\%]}\\

\midrule
iMAP             &\textbf{3.69}         & 4.87                  & \textbf{83.45}       & & 4.83                  & 6.59                  & \textbf{77.63}        \\
TANDEM           &4.26                  & \textbf{4.71}         & 81.95                & & \textbf{3.76}         & \textbf{6.11}         & 74.66                 \\
\bottomrule

\end{tabular}
\end{center}
\vspace{-0.5cm}
\end{table}

On EuRoC, we cannot compare with CNN-SLAM since the numbers on EuRoC are not provided on the paper and code is not publicly available. We further add CodeVIO~\cite{zuo2020codevio} into the evaluation on EuRoC as it is a recent dense SLAM system and it was also evaluated on EuRoC. Please note that CodeVIO uses a monocular camera and an inertial sensor for tracking, while TANDEM and other SLAM methods rely only on monocular cameras. \autoref{tab:reconst_eval_euroc} shows the evaluation results.

We further evaluate TANDEM against iMAP~\cite{Sucar:etal:iccv2021}, an RGB-D dense SLAM system that leverages deep implicit map representation. Please note two major differences between iMAP and TANDEM: on one hand, iMAP uses an RGB-D sensor while TANDEM needs only a monocular camera; on the other hand, the \ac{dnn} of iMAP does not require any pre-training and it is trained purely online with the RGB-D inputs, while the depth estimation network of TANDEM requires offline training. Since iMAP was also evaluated on Replica, we, therefore, compare TANDEM with iMAP on the two sequences of their dataset which are not included in the training set of our Replica split. We use the evaluation metrics from iMAP and show the results in \autoref{tab:imap}. In general, TANDEM achieves similar results to iMAP while using a monocular camera.

In \autoref{fig:depth_qual} we show qualitative depth maps estimated by DeepFactors, Atlas, and TANDEM. Both DeepFactors and Atlas can recover the geometry of the underlying scene well, but our method generally manages to capture more fine-scale details. We further show the complete scene reconstruction as meshes in~\autoref{fig:meshes_qual}. As DeepFactors does not generate a complete 3D model by itself, we only compare TANDEM with Atlas for this experiment. From the figure we can see that, similarly to the depth maps, TANDEM is able to reconstruct more fine-scale details than Atlas.

\section{Conclusion}\label{sec:conclusion}

We presented TANDEM, a real-time dense monocular SLAM system with a novel design that couples direct photometric visual odometry and deep multi-view stereo. In particular, we propose \ac{mvsnet} which leverages the whole keyframe window effectively and predicts high-quality depth maps. Further, the proposed dense tracking scheme bridges camera pose estimation and dense 3D reconstruction by tracking against the global 3D model created with TSDF fusion.
The quantitative and qualitative experiments show that TANDEM achieves better results than other state-of-the-art methods for both 3D reconstruction and visual odometry on synthetic and real-world data. We believe that TANDEM further bridges the gap between RGB-D mapping and monocular mapping.

\acknowledgments{We thank the anonymous reviewers for providing helpful comments that improved the paper. We express our appreciation to our colleagues, who have supported us, specifically we thank Mariia Gladkova (Technical University of Munich) and Simon Klenk (Technical University of Munich) for proof reading, as well as, Stefan Leutenegger (Technical University of Munich \& Imperial College London) and Sotiris Papatheodorou (Imperial College London) for their help with setting up a demo for the conference. We thank the authors of iMAP, specifically Edgar Sucar (Imperial College London), for providing us with evaluation data and scripts from their paper.}

\bibliography{example}

\clearpage

\appendix
\renewcommand{\thesection}{\Alph{section}}

\vspace{1cm}
{\noindent\Large\textbf{Supplementary Material}}
\section{Introduction}
In this supplementary material, we briefly introduce the volumetric mapping of TANDEM (\autoref{sec:volumetric_mapping}), discuss the initialization in \autoref{sec:init}, show further experimental results (\autoref{sec:experimental_results}), give more details regarding our experiments (\autoref{sec:experimental_details}) and, finally, provide implementation details for TANDEM in \autoref{sec:implementation_details} including the possibility to deploy TANDEM on an embedded device (cf. \autoref{sec:deployment}).
Upon publication, we will release our code for TANDEM as well as the rendered Replica sequences to facilitate the reproduction of our results.

We also urge the readers to watch the \textbf{supplementary video} which shows the real-time demos of TANDEM running on the unseen sequences.

\section{Volumetric Mapping} \label{sec:volumetric_mapping}
We use \ac{tsdf} fusion~\cite{CurlessLevoy96} to fuse the per-keyframe estimated depth maps into a globally consistent and dense 3D model. Storing the \ac{tsdf} values within a dense voxel grid has a cubic memory requirement in the spatial resolution of the grid and is thus unpractical. We employ voxel hashing~\cite{Niesner2013} to alleviate this issue. The \ac{tsdf} fusion is run on the GPU to ensure real-time performance.

For each voxel of the grid we store the estimated \ac{tsdf} value ${D_i(x) \in \R}$, the estimated RGB values ${C_i(x) \in \R^3}$, and the weight ${W_i(x) \in \R}$, where $x$ indexes the 3D location and $i$ indexes the time. After our \ac{mvsnet} has predicted a new depth map, we iterate over all voxels within the truncation distance and update their stored quantities. Let $d_{i+1}(x) \in \R$ be the projective \ac{tsdf} value of the voxel at $x$ based on the depth map from the \ac{mvsnet}. Furthermore, let $c_{i+1}(x)$ be the associated RGB value from the input image, then
\begin{align}
    D_{i+1}(x) &= \frac{W_i(x) D_i(x) + d_{i+1}}{W_i(x) + 1} \,,\\
    C_{i+1}(x) &= \frac{W_i(x) C_i(x) + c_{i+1}}{W_i(x) + 1} \,,\\
    W_{i+1}(x) &= \min( W_i(x) + 1, \,64) \,. \label{eqn:tsdf_weight}
\end{align}

The weight $W_i(x)$ is truncated at $64$ (cf.~\autoref{eqn:tsdf_weight}) to ensure that new measurements can influence the estimated quantities and avoid over-saturation.  Depth maps are rendered from the \ac{tsdf} volume using raycasting~\cite{CurlessLevoy96} and used for the proposed dense front-end tracking.

Because TANDEM is a monocular, geometry-based method the overall scale of the scene is not observable. Therefore, the voxel grid is scaled with the same scale as the visual odometry. The grid uses a voxel size of $0.01$ which roughly corresponds to $2.5 cm$ depending on the exact scale of a particular scene and run. We use a truncation distance for the \ac{tsdf} of $0.1$ which roughly corresponds to $25 cm$. The global scale ambiguity that is common to all monocular, geometry-based methods renders the voxel size scene dependent. For our experiments the scene scale does not vary considerably and therefore we use a fixed voxel size. However, when considering scenes of different extent, the voxel size and truncation distance have to be set accordingly, which is also necessary when running classical \ac{tsdf} fusion. The internal scale of TANDEM is generally such that the mean depth of the sparse points in the first frame is approximately identity.

The \ac{mvsnet} requires a minimum and maximum depth value for inference, which is usually given in the dataset. However, during the live operation of TANDEM, we do not know the relative scale between the world and our reconstruction. To facilitate the live operation of TANDEM, we choose a simple strategy: the minimum depth is set $0.01$, which is small enough to capture all objects, and the maximum depth is initially set to ten times the mean depth of the sparse points. For each following invocation of the \ac{mvsnet}, we set the maximum depth to $1.5$ times the previous maximum depth estimated by the network, which ensures that the whole scene is covered while providing good depth resolution. We found this simple scheme to work well in our experiments.

\section{Initialization}\label{sec:init}
The proposed \ac{mvsnet} operates on a keyframe window and thus the tracking is initialized the same way as in DSO~\cite{Engel2018dso} with non-linear optimization on poses and the sparse depth. The \ac{tsdf} volume is initialized to represent empty space, i.e. with zero weights $W_0(x) = 0 \; \forall x$. Because we use voxel hashing \cite{Niesner2013} to represent the \ac{tsdf} volume, empty space can be represented very efficiently by not allocating any voxel blocks. After the first dense depth map is predicted and integrated into the \ac{tsdf} volume, TANDEM uses the rendered nearly dense depth maps for tracking as can be seen from \autoref{fig:dense_tracking1_of2}.

\section{Further Experimental Results} \label{sec:experimental_results}
\begin{figure}[tb]
  \centering
  \includegraphics[width=3.4in]{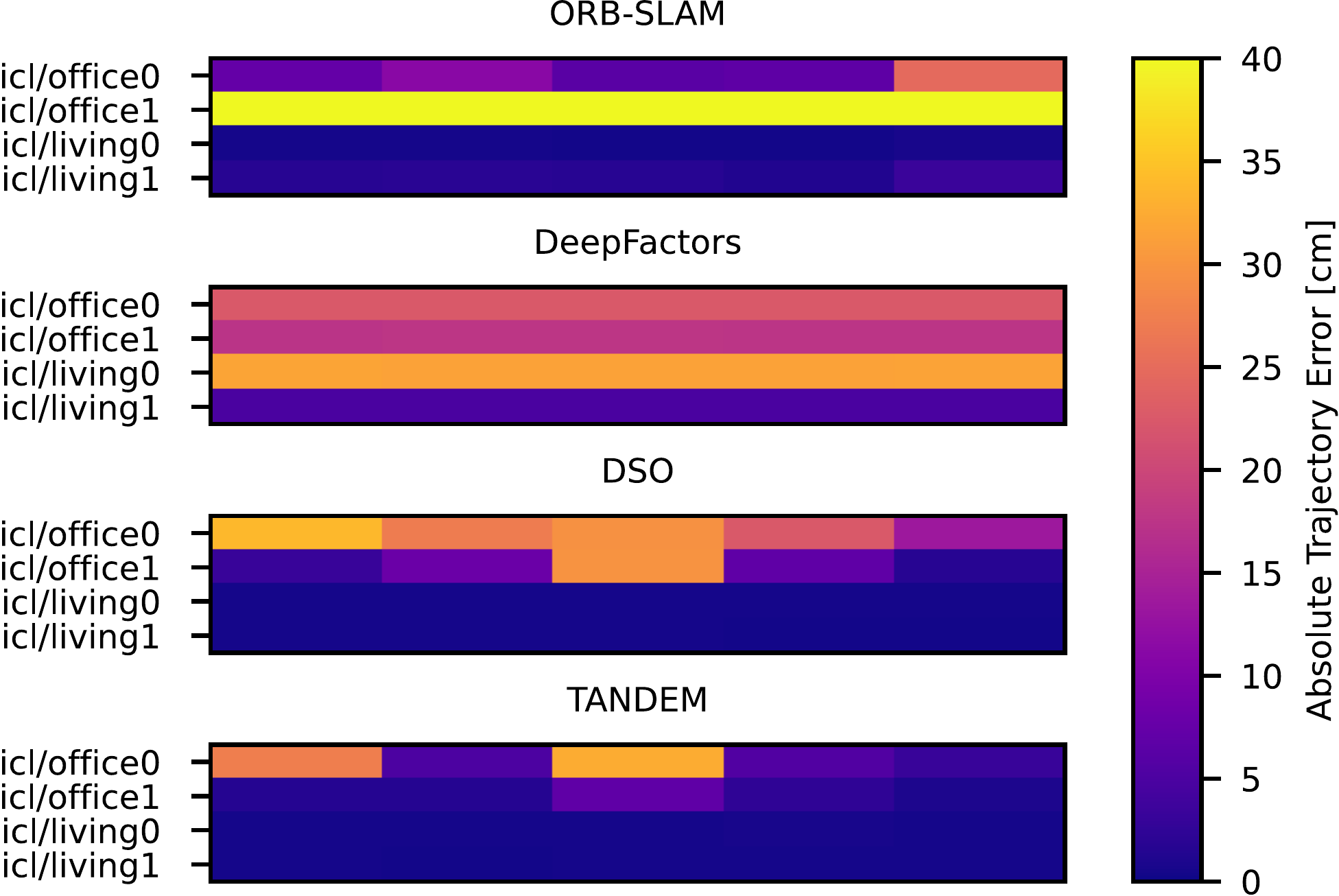}
  \caption{Trajectory evaluation on the ICL-NUIM dataset. Because all systems use monocular images we perform $\text{Sim}(3)$ alignment w.r.t. the ground truth trajectories. We show the color-coded absolute trajectory error (ATE) for five runs. The results show that our dense tracking improves the robustness over DSO. DeepFactors produces nearly identical results for all five runs in this experiment but shows a higher median error.}
  \label{fig:trajectory_eval}
\end{figure}

\begin{figure}[tb]
  \centering
    \includegraphics[width=3.4in]{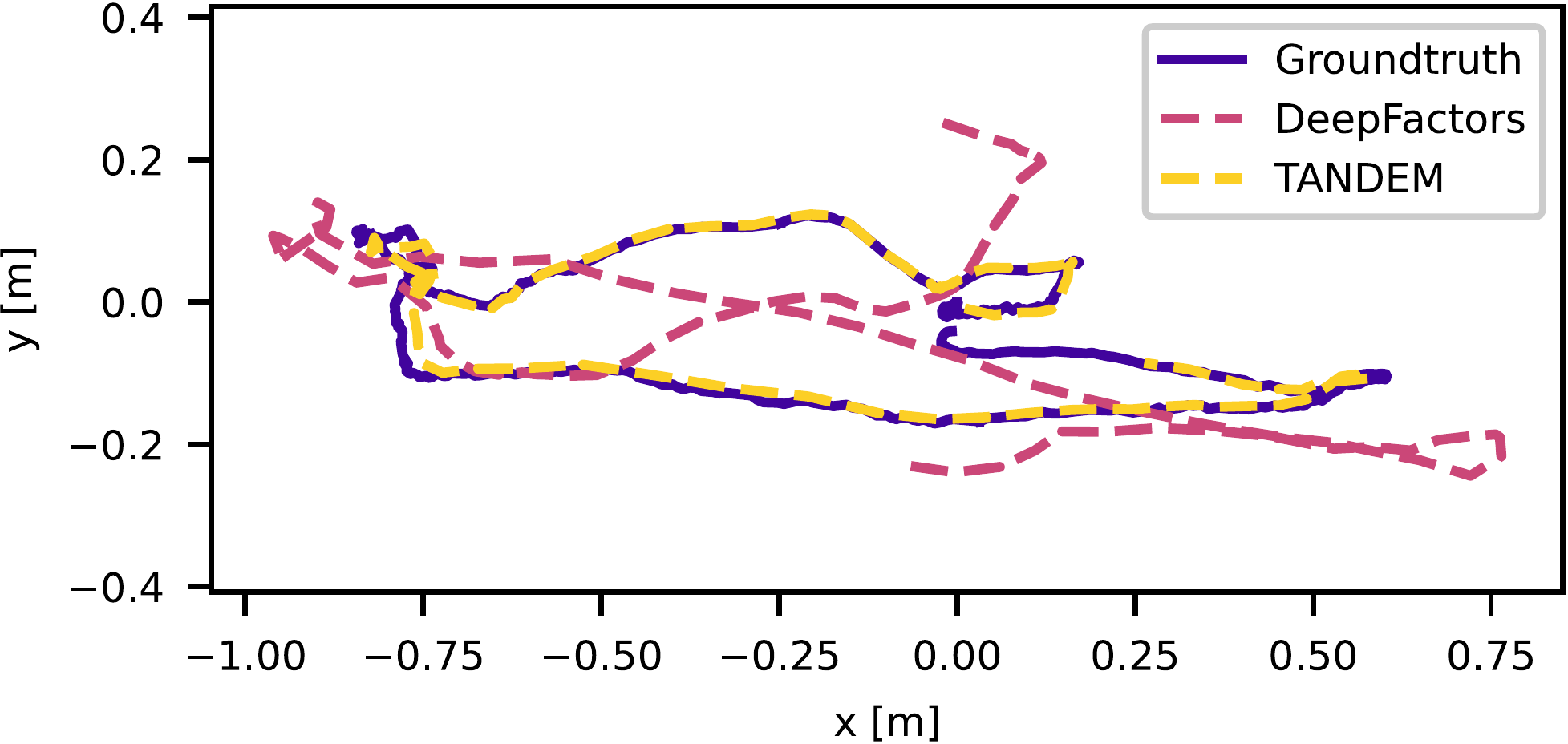}
\caption{Estimated trajectories of DeepFactors and TANDEM on icl/office1. For both methods we select the run with the median error (cf. \autoref{fig:trajectory_eval}).}
  \label{fig:traj_comp}
\end{figure}

\begin{figure}[tb]
  \centering
  \includegraphics[width=\textwidth]{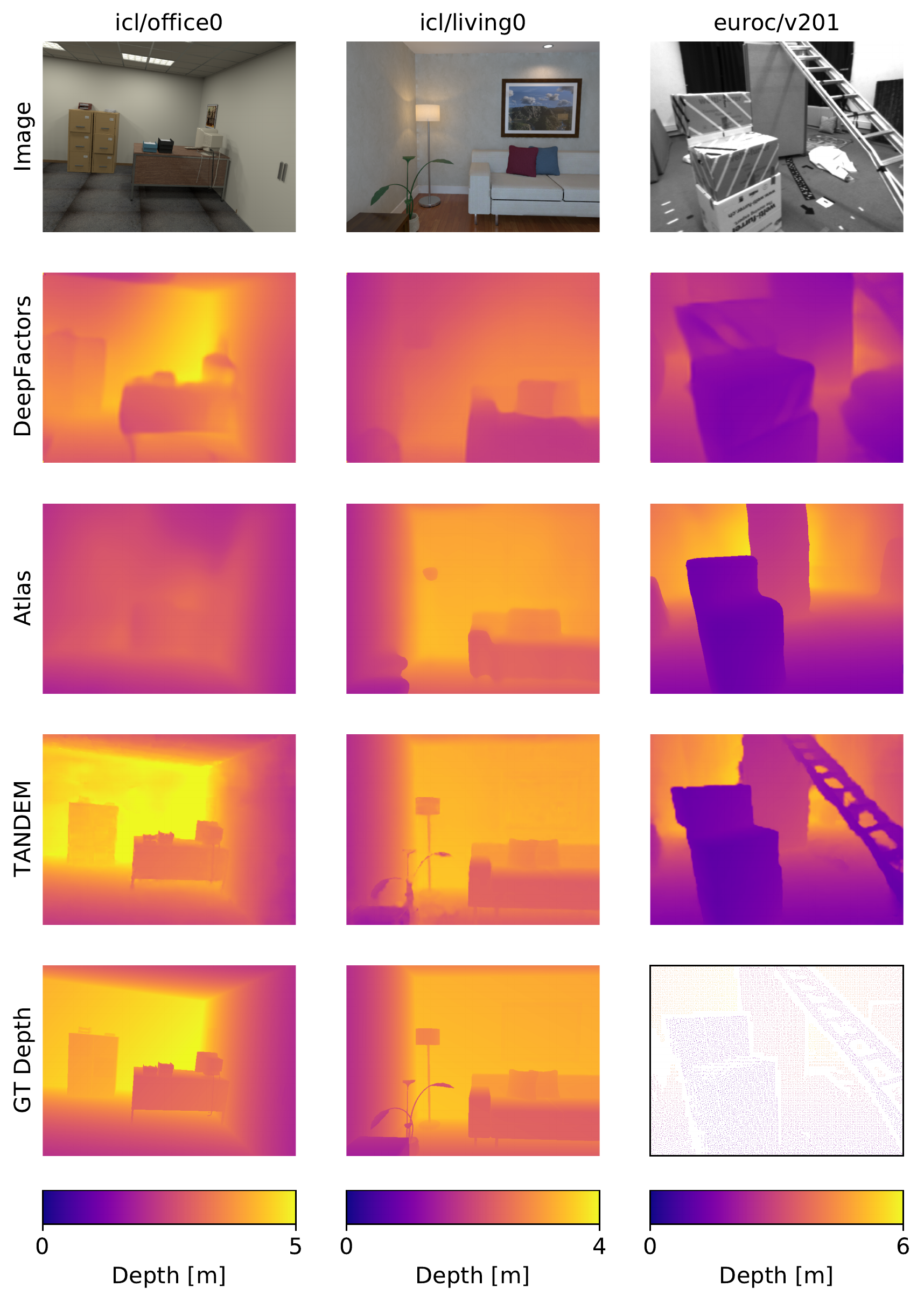}
  \caption{Depth comparison for DeepFactors~\cite{deep_factors}, Atlas~\cite{murez2020atlas}, and TANDEM on unseen sequences. TANDEM produces finer-scale details, e.g. the plant in the second column, or the ladder in the third column. For EuRoC only sparse ground-truth depth is available. This is a high-resolution version of Figure 3 from the main paper. In \autoref{fig:qual_depth_hr2} we show further qualitative results.}
  \label{fig:qual_depth_hr1}
\end{figure}

\begin{figure}[tb]
  \centering
  \includegraphics[width=\textwidth]{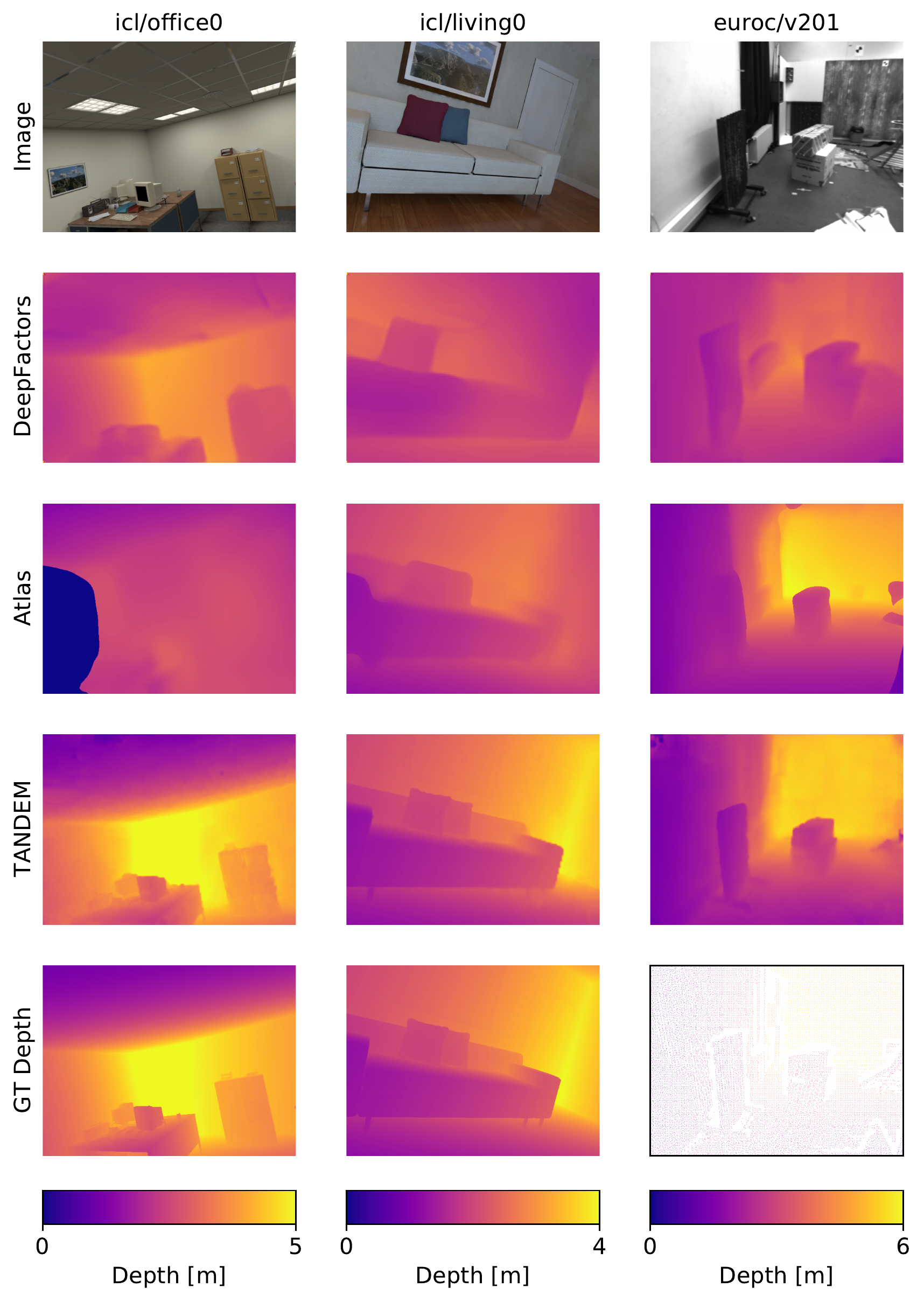}
  \caption{Further depth comparison for DeepFactors~\cite{deep_factors}, Atlas~\cite{murez2020atlas}, and TANDEM on unseen sequences. TANDEM produces finer-scale details, e.g. the computers in the first column, the pillows in the second column, or the radiator in the third column. However, TANDEM can produce outliers due to the cost volume-based formulation (cf. third column upper left corner of the image). For Atlas, the reconstruction of office0 is too small and thus the rendered depth has an invalid region (the blue blob in bottom left). For EuRoC only sparse ground-truth depth is available.}
  \label{fig:qual_depth_hr2}
\end{figure}

\begin{figure}[tb]
  \centering
  \includegraphics[width=0.94\textwidth]{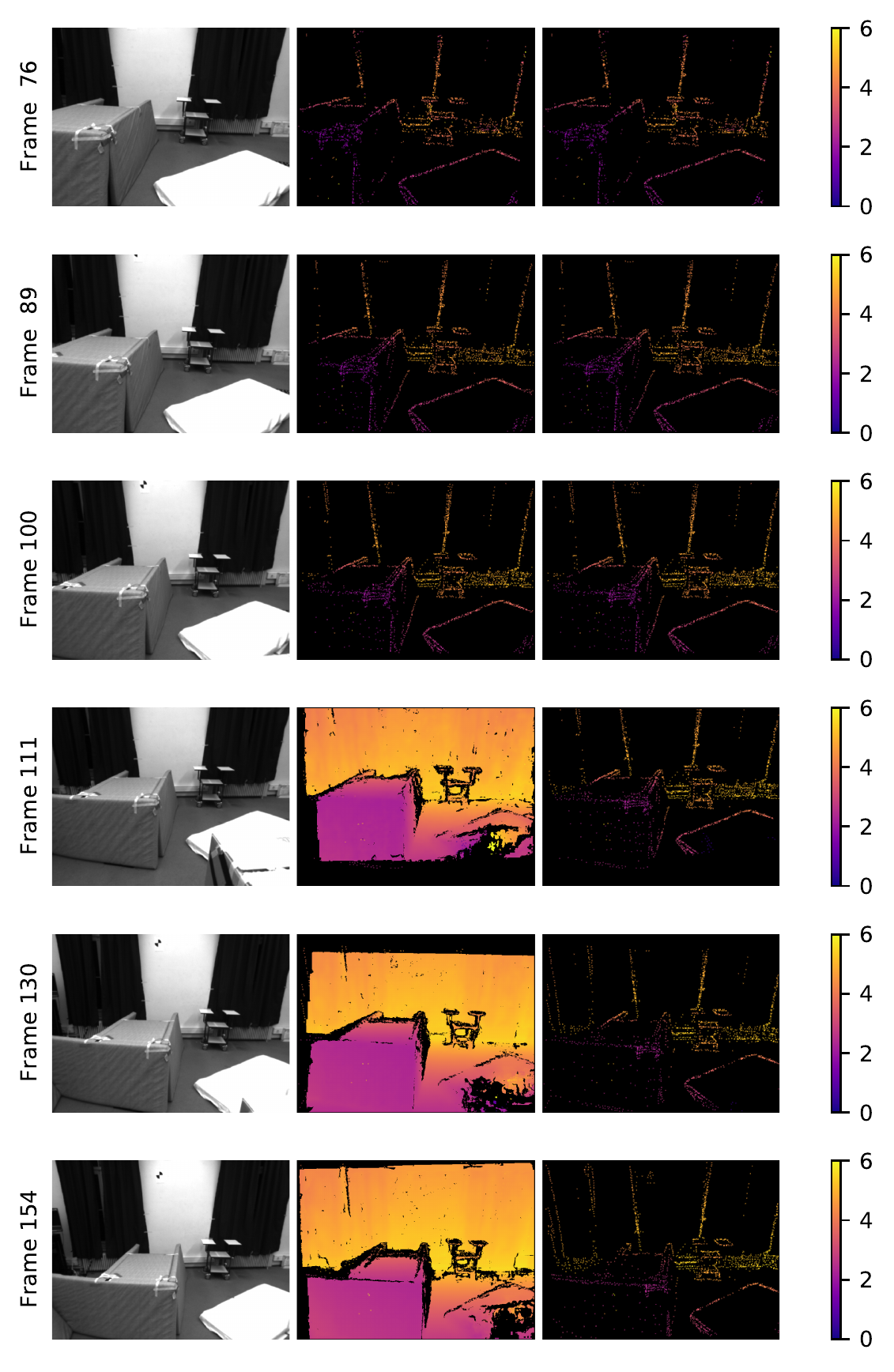}
  \caption{Tracking Depth Maps for EuRoC/V201. For every second keyframe we show the image (\textit{left}), the depth map used for tracking by TANDEM (\textit{middle}), and the sparse depth map that would be used without the dense tracking (\textit{right}). For the first few keyframes no dense depth is available and thus TANDEM uses the sparse depth map for tracking. Note that, as in DSO, the sparse depth maps are slightly dilated before they are used for tracking. Further frames are shown in \autoref{fig:dense_tracking2_of2}.}
  \label{fig:dense_tracking1_of2}
\end{figure}

\begin{figure}[tb]
  \centering
  \includegraphics[width=0.94\textwidth]{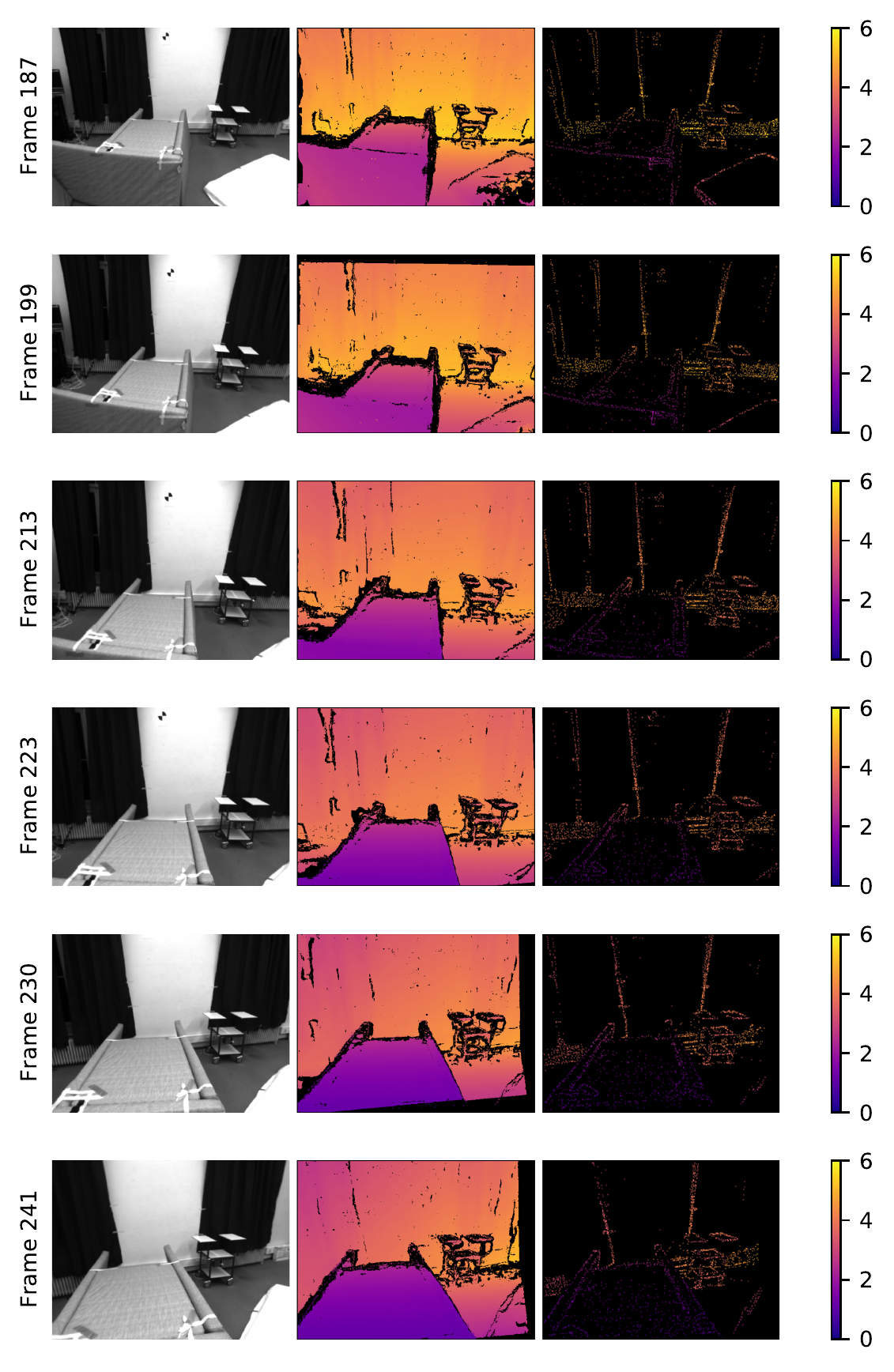}
  \caption{Tracking Depth Maps for EuRoC/V201. For every second keyframe we show the image (\textit{left}), the depth map used for tracking by TANDEM (\textit{middle}), and the sparse depth map that would be used without the dense tracking (\textit{right}). The nearly dense depth map represents the scene well and thus the dense tracking can give more accurate results than the sparse tracking \cite{DBLP:journals/ivc/StrasdatMD12,DBLP:phd/ethos/Newcombe12}. The dense depth map also incorporates the sparse depth values which can be seen at occlusion boundaries. Note that, as in DSO, the sparse depth maps are slightly dilated. Previous frames are shown in \autoref{fig:dense_tracking1_of2}.}
  \label{fig:dense_tracking2_of2}
\end{figure}

We show the trajectory evaluation on ICL-NUIM in \autoref{fig:trajectory_eval} and \autoref{tab:trajectory_icl}. Similar to the evaluation on the EuRoC dataset in the main paper, we see that the proposed TANDEM shows better results than DSO and DeepFactors~\cite{deep_factors}. Furthermore, TANDEM performs favourably in comparison to the tracking component of DeepTAM~\cite{zhou2018deeptam} that uses ground truth depth maps. We compare the trajectories for DeepFactors and TANDEM on the office1 sequence of the ICL-NUIM dataset in \autoref{fig:traj_comp}.
In \autoref{fig:qual_depth_hr1} we reproduce Figure 3 from the main paper in higher resolution to enable closer inspection of the generated depth maps. Furthermore, we show additional qualitative depth map comparisons in \autoref{fig:qual_depth_hr2}.
In \autoref{sec:failure_cases} we discuss failure cases and challenges for TANDEM.
The nearly dense depth maps used for tracking by TANDEM and the sparse depth maps from the photometric bundle adjustment are shown in \autoref{fig:dense_tracking1_of2} and \autoref{fig:dense_tracking2_of2}.

\subsection{Failure Cases and Challenges} \label{sec:failure_cases}
Dense monocular SLAM is a challenging problem~\cite{DBLP:phd/ethos/Newcombe12} and most systems fail in certain scenarios or perform worse given certain challenges. Any system that involves deep learning has to consider the generalization capability of neural network since, for SLAM, the training and testing datasets will practically always differ. Pure rotational motion is a known challenge for both visual SLAM and 3D reconstruction due to the unobservability of depth. Dynamic scenes, rolling shutter, and geometric errors are well known failure cases in SLAM if they are not modelled explicitly. While some of the aforementioned failure cases and challenges can be overcome by explicit modelling or through employing different techniques for tracking or mapping, this often leads to an increase in runtime and system complexity. In the following, we describe the weaknesses and the failure cases of the proposed system in order to facilitate further research in this direction:

\paragraph{Neural Network Generalization}
Generalization of deep neural networks is well known to pose challenges \cite{DBLP:journals/corr/abs-1710-05468} and a general solution has yet to be proposed. For deep \ac{mvs} multiple remedies have been proposed, including diverse training datasets \cite{DBLP:conf/cvpr/0008LLZRZFQ20}, specific architecture choices \cite{DBLP:conf/3dim/CaiPMM20,zhang2019domaininvariant}, and self-supervised adaption \cite{DBLP:conf/aaai/XuZ0KW21,poggi2021continual}. While our \ac{mvsnet} already shows good generalization capabilities as it was trained on either the ScanNet or the Replica dataset and generalizes to the ICL-NUIM and EuRoC datasets, the error metrics are better for the training datasets. Although initial experiments with specific architectures as well as self-supervised adaption didn't show convincing results for our training and testing datasets, we consider this a good direction for future research. For practical applications we consider the utilization of diverse training data together with re-training for new scenarios the most promising direction.

\paragraph{Pure Rotational Motion}
Pure rotational motion is critical because the depth of a point cannot be inferred, which leads to problems for tracking as well as depth estimation using \ac{mvs}. While some purely rotational motion can be handled by TANDEM if the system has been initialized well, rotation-only motion during initialization will lead to failure in tracking. We experimentally validate this using the sequence office0 from the ICL-NUIM dataset. This sequence contains highly rotational motion and thus the results of TANDEM and DSO are worse in comparison to the office1 sequence or the living room sequences (cf.~\autoref{tab:trajectory_icl}). We investigate the effect of rotational motion during initialization by starting the office0 sequence at frame $300$, which is followed by a rotation along the ceiling of the office. Although the subsequence is shorter than the full sequence, the mean RMSE of the absolute pose error for TANDEM increases from $0.056 \,m$ to $0.354 \,m$, which shows that rotational motion during initialization remains an open challenge.

\paragraph{Dynamic Scenes}
While the photometric bundle adjustment of TANDEM can implicitly handle dynamic objects if they constitute a relatively small fraction of the scene, the deep \ac{mvs} fully relies on a static scene. Such scenarios are favourable for mono-to-depth-based methods as they are independent of the scene dynamics when predicting the depth. On the other hand, dynamic objects can be explicitly modelled within the cost volume \cite{wimbauer2021monorec}.

\paragraph{Rolling Shutter}
The rolling shutter effect is known to be problematic for direct SLAM because not all pixels correspond to the same time and thus the same pose \cite{DBLP:conf/eccv/SchubertDUSC18}. To investigate the effect of rolling shutter on TANDEM we use the ICL-NUIM living room sequences with synthetic rolling shutter proposed in \cite{DBLP:conf/iccv/KerlSC15}. For the sequences living0 and living1 the mean ATE RMSE increases from $0.006 \,m$ to $0.018 \,m$ and from $0.005 \,m$ to $0.074 \,m$, respectively.

\paragraph{Geometric Error}
Photometric methods are susceptible to geometric errors, e.g. from an inaccurate camera calibration, because these geometric errors are, in contrast to feature-based methods, not modeled explicitly. We simulate geometric errors by using an incorrect calibration for the ICL-NUIM living room sequences, specifically we add $5\,px$ to $f_x, f_y, c_x$, and $c_y$.  For the sequences living0 and living1 the mean ATE RMSE increases from $0.006 \,m$ to $0.041 \,m$ and from $0.005 \,m$ to $0.026 \,m$, respectively.

\begin{table}[tb]
\small
\caption{Pose evaluation on ICL-NUIM~\cite{HandaWMD14}. All the methods except DeepTAM are $\text{Sim} (3)$ aligned w.r.t. the ground-truth trajectories. The mean absolute pose errors in meters and the standard deviations over five runs are shown. \textsuperscript{\textdagger}For DeepTAM the official implementation contains only a pure tracking mode that requires ground truth depth maps for tracking, therefore, $\text{SE} (3)$ alignment is performed because the scale is observable through the depth maps. Note that we show the result of a single run for DeepTAM since the results are consistent across multiple runs.}
\label{tab:trajectory_icl}
\begin{center}
\begin{tabular}{p{1.7cm} c c c c c}
    \toprule
    Sequence       &
    \makecell{DeepTAM\textsuperscript{\textdagger}~\cite{zhou2018deeptam}}  &
    \makecell{DeepFactors~\cite{deep_factors}}  &
    \makecell{DSO~\cite{Engel2018dso}} &
    \makecell{TANDEM} \\
    \midrule
icl/office0    &  \underline{0.086}  (-) &                0.226   ($\pm$ 0.000)  &                0.270   ($\pm$ 0.070)  &    \textbf{0.056} ($\pm$ 0.126) \\
icl/office1    &  \underline{0.030}  (-) &                0.175   ($\pm$ 0.001)  &                0.069   ($\pm$ 0.102)  &    \textbf{0.017} ($\pm$ 0.021) \\
icl/living0    &                 0.025    (-) &                0.315   ($\pm$ 0.001)  &    \textbf{0.006} ($\pm$ 0.000)  & \underline{0.006} ($\pm$ 0.000) \\
icl/living1    &                 0.034    (-) &                0.049   ($\pm$ 0.000)  & \underline{0.005} ($\pm$ 0.000)  &    \textbf{0.005} ($\pm$ 0.000) \\
\bottomrule
\end{tabular}
\end{center}
\vspace{-0.5cm}
\end{table}

\section{Experimental Details} \label{sec:experimental_details}
\subsection{Poses and Alignment}
For the ablation study, we use ground-truth poses and a fixed set of keyframe windows to enable a fair and accurate comparison.

For the depth evaluation on ICL-NUIM and EuRoC, we run our visual odometry once to generate poses as well as keyframe windows. We use the same poses and keyframe windows to evaluate Cas-MVSNet, Ours (ScanNet), and Ours (Replica) to enable a fair and accurate comparison. Evaluating based on the full system would introduce significant non-determinism into the results. Because TANDEM is a monocular method, we use the scale estimated from the Sim(3) alignment of the trajectory to scale the depth maps. Since DeepFactors does not estimate very accurate poses on the EuRoC dataset (cf. Table 4 in the main paper), we scale align each depth map individually using the median scale between the ground-truth depth map and the predicted depth map. This procedure potentially overestimates the accuracy of DeepFactors, which is important to note when comparing with other methods.

For the comparison to iMAP, we use the Sim(3) alignment of the trajectory of \mbox{TANDEM} and the ground truth trajectory to transform the mesh into the same coordinate frame as the reference.

For the comparison on EuRoC we rectify the images s.t. the resulting images have resolution ${640 \times 480}$ and evaluate all methods except CodeVIO on the same rectified images for a fair comparison. For CodeVIO no public source code is available and we thus use the numbers published by the authors. Using a different rectification protocol can result in different results, which is often not obvious from the paper.

\subsection{Depth Evaluation Metrics}
For the following we will use $y_i^*$ to denote the ground-truth depth value in meters and $y_i$ to denote the corresponding predicted depth value. All metrics are computed per image first and then the average is taken over all images within the sequence. We let the index $i = 1, \dots, N$ enumerate all pixels with valid ground-truth depth for a given image. Note that thus there holds $y_i^* > 0, \, \forall i$.

The paper uses inconsistent metrics, e.g. $a_1$ and $d_1$, because prior works, e.g. DeepFactors and CodeVIO, used inconsistent metrics. Additionally, for some prior works, e.g. CodeVIO and CNN-SLAM, the source code is not available and therefore we cannot evaluate all methods using one consistent metric.

For the comparison on ICL-NUIM we follow DeepFactors and use the $a_1$ metric
\begin{equation}
    a_1 = \frac{100}{N} \sum_{i=1}^N \ind(|y_i - y_i^*|/y_i^* < 0.1) \,,
\end{equation}
where $\ind(\cdot)$ is the indicator function. The $a_1$  metric gives the percentage of pixels for which the estimated depth falls within $10 \%$ of the ground-truth depth.

For the ablation study, we follow DeepFactors and use the $a_1$ metric. We found that this metric is saturated because all configurations reach more than $98 \%$ and thus also show the stricter $a_2$ and $a_3$ metrics defined by
\begin{align}
    a_2 &= \frac{100}{N} \sum_{i=1}^N \ind(|y_i - y_i^*|/y_i^* < 0.01) \,,\\
    a_3 &= \frac{100}{N} \sum_{i=1}^N \ind(|y_i - y_i^*|/y_i^* < 0.001) \,.
\end{align}
The $a_2$ and $a_3$  metrics give the percentage of pixels for which the estimated depth falls within $1\%$ and $0.1\%$ of the ground truth depth.
Additionally, we show the mean absolute error in centimeters
\begin{equation}
    Abs = \frac{100}{N} \sum_{i=1}^N |y_i - y_i^*| \,.
\end{equation}

For the comparison on EuRoC we follow CodeVIO and use the $d_1$ metric as introduced by Eigen et al.~\cite{eigen_split}
\begin{equation}
    d_1 = \frac{100}{N} \sum_{i=1}^N \ind\Big(\max(\frac{y_i}{y_i^*}, \frac{y_i^*}{y_i}) < 1.25 \Big) \,.
\end{equation}

\section{Implementation Details} \label{sec:implementation_details}
\subsection{CVA-MVSNet}
We list all the hyperparameters and their corresponding values in \autoref{tab:hyperparameter}.

\textbf{Architecture.} The proposed \ac{mvsnet} consists of three trainable components: the feature extraction network~(cf.~\autoref{tab:feature_net_arch}), the view aggregation network~(cf.~\autoref{tab:va_arch}), and the cost regularization network~(cf.~\autoref{tab:cost_net_arch}).
The feature extraction network takes as input each frame $I_i$ and outputs the corresponding multi-scale feature maps $\mathbf{F}_i^s$, which are denoted by  \texttt{out.stage1}, \texttt{out.stage2}, \texttt{out.stage3} in \autoref{tab:feature_net_arch}. The weights are shared for all frames.
The view aggregation network takes as inputs the single frame cost volumes $(\featVol_i^s - \featVol_j^s)^2$ and outputs the aggregation weights $\vaW_i^s$. The weights are shared for all frames but are not shared for the three stages.
The cost regularization network takes as input the aggregated cost volume
\begin{equation*}
    \costVol^s = \frac{\sum_{i = 1, i \neq j}^n (1 + \vaW_i^s) \, \odot \, (\featVol_i^s - \featVol_j^s)^2}{n - 1} \, ,
\end{equation*}
and outputs the probability volume $\probVol^s$. There is a single network for all frames and the weights are not shared for the three stages.

\begin{table}[tb]
\small
\caption{Feature extraction network. Layers are 2D convolutions or nearest neighbor interpolation with a scale factor $2$, which we denote by $\uparrow$. For all layers we show the input and the channels~(\textbf{chns}). For convolutions, we additionally show the kernel size~(\textbf{k}), the stride~(\textbf{s}), the padding~(\textbf{p}), if the layer uses batch normalization~(\textbf{bn}), and the activation function~(\textbf{act}). Only the layers \texttt{skip.stage2} and \texttt{skip.stage3} have a bias term. We use batch normalization before the activation function.}
\label{tab:feature_net_arch}
\begin{center}
\begin{tabular}{l c c c c c c c}
    \toprule
    \textbf{Layer} & 
    \textbf{input} & 
    \textbf{k} &
    \textbf{s} &
    \textbf{p} &
    \textbf{chns} &
    \textbf{bn} & 
    \textbf{act} \\
    \midrule
    \texttt{conv0.0} & \texttt{image} & 3 & 1 & 1& $\phantom{0}3 \to \phantom{0}8$& \checkmark & ReLU \\
    \texttt{conv0.1} & \texttt{conv0.0} & 3 & 1 & 1& $\phantom{0}8 \to \phantom{0}8$& \checkmark & ReLU \\
    \midrule
    \texttt{conv1.0} & \texttt{conv0.1} & 5 & 2 & 2 & $\phantom{0}8 \to 16$& \checkmark & ReLU \\
    \texttt{conv1.1} & \texttt{conv1.0} & 3 & 1 & 1 & $16 \to 16$& \checkmark & ReLU \\
    \texttt{conv1.2} & \texttt{conv1.1} & 3 & 1 & 1 & $16 \to 16$& \checkmark & ReLU \\
    \midrule
    \texttt{conv2.0} & \texttt{conv1.2} & 5 & 2 & 2 & $16 \to 32$& \checkmark & ReLU \\
    \texttt{conv2.1} & \texttt{conv2.0} & 3 & 1 & 1 & $32 \to 32$& \checkmark & ReLU \\
    \texttt{conv2.2} & \texttt{conv2.1} & 3 & 1 & 1 & $32 \to 32$& \checkmark & ReLU \\
    \midrule
    \texttt{out.stage1} & \texttt{conv2.2} & 1 & 1 & 0 & $32 \to 32$ & & \\
    \midrule
    \texttt{skip.stage2} & \texttt{conv1.2} & 1 & 1 & 0 & $16 \to 32$ &  &  \\
    \texttt{inter.stage2} & $\uparrow\texttt{conv2.2} + \texttt{skip.stage2}$ &  &  &  & $32 \to 32$ &  &  \\
    \texttt{out.stage2} & \texttt{inter.stage2} & 3 & 1 & 1 & $32 \to 16$ &  &  \\
    \midrule
    \texttt{skip.stage3} & \texttt{conv0.1} & 1 & 1 & 0 & $\phantom{0}8 \to 32$ &  &  \\
    \texttt{inter.stage3} & $\uparrow\texttt{inter.stage2} + \texttt{skip.stage3}$ &  &  &  & $32 \to 32$ &  &  \\
    \texttt{out.stage3} & \texttt{inter.stage3} & 3 & 1 & 1 & $32 \to \phantom{0}8$ &  &  \\
    \bottomrule
\end{tabular}
\end{center}
\nocite{DBLP:conf/icml/IoffeS15}
\nocite{fukushima1982neocognitron}
\nocite{nair2010rectified}
\end{table}
\begin{table}[tb]
\small
\caption{View aggregation network. All layers are 3D convolutions. We show the input, the kernel size~(\textbf{k}), the stride~(\textbf{s}), the padding~(\textbf{p}), the channels~(\textbf{chns}), if the layer uses batch normalization~(\textbf{bn}), and the activation function~(\textbf{act}). All layers have a bias term. We use batch normalization before the activation function. The single frame cost volumes~(\texttt{single frame cost volume}) have $32$ channels for stage 1, $16$ channels for stage 2, and $8$ channels for stage 3.}
\label{tab:va_arch}
\begin{center}
\begin{tabular}{l c c c c c c c c}
    \toprule
    \textbf{Layer} & 
    \textbf{input} & 
    \textbf{k} &
    \textbf{s} &
    \textbf{p} &
    \textbf{chns} &
    \textbf{bn} & 
    \textbf{act} \\
    \midrule
    \texttt{conv0} & \makecell{\texttt{single frame}\\\texttt{cost volume}} & 1 & 1 & 0 & $(32, 16, 8) \to \phantom{0}1$& \checkmark & ReLU \\
    \midrule
    \texttt{conv1} & \texttt{conv0} & 1 & 1 & 0 & $\phantom{0}1 \to \phantom{0}1$& \checkmark & ReLU \\
    \bottomrule
\end{tabular}
\end{center}
\end{table}
\begin{table}[tb]
\small
\caption{Cost regularization network. All layers are either 3D convolutions~(\texttt{conv}) or 3D transposed convolutions~(\texttt{convT}). For all layers, we show the input, the kernel size~(\textbf{k}), the stride~(\textbf{s}), the padding~(\textbf{p}), the channels~(\textbf{chns}), if the layer uses batch normalization~(\textbf{bn}), and the activation function~(\textbf{act}). For transposed convolutions we show the output padding~(\textbf{op}). We use batch normalization before the activation function. No layer has a bias term. The stride $S$ (cf. \texttt{conv5} and \texttt{convT7}) is $2$ for the first stage of \ac{mvsnet} and $(1,2,2)$ for stages 2 and 3 because of the reduced number of depth planes. For the same reason, the output padding $P$ (cf. \texttt{convT7}) is $1$ for stage 1 and $(0, 1, 1)$ for stages 2 and 3. The cost volume~(\texttt{aggregated cost volume}) has $32$ channels for stage 1, $16$ channels for stage 2, and $8$ channels for stage 3.}
\label{tab:cost_net_arch}
\begin{center}
\begin{tabular}{l c c c c c c c c}
    \toprule
    \textbf{Layer} & 
    \textbf{input} & 
    \textbf{k} &
    \textbf{s} &
    \textbf{p} &
    \textbf{op} &
    \textbf{chns} &
    \textbf{bn} & 
    \textbf{act} \\
    \midrule
    \texttt{conv0} & \makecell{\texttt{aggregated}\\\texttt{cost volume}} & 3 & 1 & 1 & & $(32, 16, 8) \to \phantom{0}8$& \checkmark & ReLU \\
    \midrule
    \texttt{conv1} & \texttt{conv0} & 3 & 2 & 1 & & $\phantom{0}8 \to 16$& \checkmark & ReLU \\
    \texttt{conv2} & \texttt{conv1} & 3 & 1 & 1 & & $16 \to 16$& \checkmark & ReLU \\
    \midrule
    \texttt{conv3} & \texttt{conv2} & 3 & 2 & 1 & & $16 \to 32$& \checkmark & ReLU \\
    \texttt{conv4} & \texttt{conv3} & 3 & 1 & 1 & & $32 \to 32$& \checkmark & ReLU \\
    \midrule
    \texttt{conv5} & \texttt{conv4} & 3 & $S$ & 1 & & $32 \to 64$& \checkmark & ReLU \\
    \texttt{conv6} & \texttt{conv5} & 3 & 1 & 1 & & $64 \to 64$& \checkmark & ReLU \\
    \midrule
    \texttt{convT7} & \texttt{conv6} & 3 & $S$ & 1 & $P$ &$64 \to 32$& \checkmark & ReLU \\
    \midrule
    \texttt{convT8} & \texttt{conv4} + \texttt{convT7} & 3 & 2 & 1 & 1 &$32 \to 16$& \checkmark & ReLU \\
    \midrule
    \texttt{convT9} & \texttt{conv2} + \texttt{convT8} & 3 & 2 & 1 & 1 & $16 \to \phantom{0}8$& \checkmark & ReLU \\
    \midrule
    \texttt{prob} & \texttt{conv0} + \texttt{convT9} & 3 & 1 & 1 & & $\phantom{0}8 \to \phantom{0}1$&  & $\operatorname{Softmax}_{\text{depth}}$ \\
    \bottomrule
\end{tabular}
\end{center}
\end{table}
\begin{table}[tb]
\small
\caption{Hyperparameter Table.}
\label{tab:hyperparameter}
\begin{center}
\renewcommand{\arraystretch}{1.5}
\begin{tabular}{p{3.2cm} c p{6.5cm}}
    \toprule
    &
    \makecell{Value} &
    \makecell{Description}\\
    \midrule
    \texttt{min depth} & $0.01$  & Minimal depth value in meters used for generating the depth planes for our \ac{mvsnet}. \\
    \texttt{max depth} & $10.0$  & Maximal depth value in meters used for generating the depth planes for our \ac{mvsnet}. \\
    \texttt{depth planes} & $(48, 4, 4)$  & Number of depth planes for each stage. \\
    \texttt{depth intervals} & $(0.213, 0.106, 0.053)$  & The distance in meters between two depth planes for each stage. For the first stage, the planes are evenly spaced between the minimum and maximum depth. For stages 2 and 3 the interval is divided by 2 and 4, respectively. \\
    optimizer & Adam & We use the default parameters: $(\beta_1, \beta_2) = (0.9, 0.999)$ and $\epsilon = 10^{-8}$. \\
    learning rate & $0.004$ &  Learning rate at the start of the schedule. \\
    learning rate schedule & linear decay &  Linear decay from \texttt{lr} to $\texttt{lr} / 100$. \\
    epochs & 50 & The number of training epochs on the Replica dataset. \\
    batch size & $4 \times 2$ & We train with $4$ GPUs with batch size $2$ on each GPU without synchronized batch norm. \\
    BN momentum & $0.1$ & Batch normalization momentum. \\
    image size & $640 \times 480$ & The size of images and depth maps on the finest scale, i.e. stage three.\\
    \bottomrule
\end{tabular}
\end{center}
\nocite{kingma_adam_2015}
\end{table}

\subsection{Runtime}
\begin{table}[tb]
\small
\caption{Timing results for TANDEM on EuRoC/V101. We show the averaged per-frame times and the averaged per-keyframe times and timed 2685 single frames for which 722 keyframes were created. Processing one frame takes $47$ ms, which gives a throughput of ca. $21$ FPS.
The \ac{mvsnet}, \ac{tsdf} fusion, and Bundle Adjustment are run only for keyframes and thus we show the time taken per keyframe as well as the average time taken per frame, which is ca. four times lower due to the ratio of frames and keyframes.
Note that the \ac{mvsnet} per-keyframe time is lower than what we reported in the main paper (cf. Table 1) because we use asynchronicity and parallelism between CPU and GPU to ensure real-time performance.}
\label{tab:timing_v101}
\begin{center}
\begin{tabular}{l c @{\hspace{0.5cm}} c}
    \toprule
    & 
    \textbf{Per Frame [ms]} & 
    \textbf{Per Keyframe [ms]} \\
    Number & 2685 & 722 \\
    \toprule
    \ac{mvsnet} & \multirow{ 2}{*}{22.6} & 53.0 \\
    \ac{tsdf} Fusion &  & 29.3 \\
    \midrule
    Coarse Tracking & 10.5 & \\
    Bundle Adjustment & 13.9 & 45.2 \\
    \midrule \midrule
    Sum & 47.0 &  \\
    \bottomrule
\end{tabular}
\end{center}
\end{table}
We report the mean runtimes for the single components of TANDEM in \autoref{tab:timing_v101}. Overall, TANDEM requires an average of 47 ms of processing time per frame, which gives a throughput of ca. 21 FPS. Additionally, \autoref{tab:timing_v101} shows that using asynchronicity and parallelism between CPU and GPU is necessary to achieve real-time capability. 

\subsection{Deployment on an Embedded Device} \label{sec:deployment}
For a SLAM system in the context of mobile robotics the deployment to an embedded platform is necessary for operation in the real world, if no uninterrupted connection to a server is available to offload computation. While we consider the actual deployment to an embedded system outside the scope of this research work, we show in the following that such a deployment is possible while maintaining accurate results.

An embedded system can benefit from software optimizations such as 16-bit float inference or NVIDIA TensorRT~\cite{tensorRT}, which can bring speedups of up to $2\times$ and $2.5\times$, respectively \cite{tensor_rt_gtc2020}. However, we consider these engineering-focused optimizations outside the scope of this work.

The inference time for our \ac{mvsnet} given in the main paper is 158 ms and decreases to 133 ms by switching from PyTorch 1.5 to PyTorch 1.9. The 2D feature extraction network is run on all images from the keyframe window, while for all but one image the feature maps have been computed before. This fact can be used to save computational cost, however, to simplify the implementation, we did not use this optimization for the timings in Table 1. If the feature extraction network is run for only one image, the overall inference time decreases from 133 ms to 118 ms.

For the EuRoC experiment in the main paper we use images of size $640 \times 480$, which is relatively high in comparison to the image size used by DeepFactors of $256 \times 192$. Using this smaller image size decreases the inference time from 118 ms to 26.5 ms, a 77.5\% relative improvement. The model trained on $640\times480$ images can be used for any resolution due to the fully-convolutional and geometry-based \ac{mvsnet} and achieves an absolute error of 6.52 cm on the Replica validation set in comparison to 2.33 cm for the full resolution. However, when evaluating on EuRoC using the lower resolution results in an average accuracy ($d_1$) of 91.43\% in comparison to 94.40\%. The much smaller difference is due to the domain shift and additionally due to the relative laxness of the $d_1$ metric. Overall, the lower-resolution model is computationally much more efficient at a reasonable accuracy decrease.

Finally, TANDEM produces new keyframes often and marginalizes them early, which has been shown to aid tracking \cite{Engel2018dso,MurArtal2015}. A new keyframe is created roughly every 5 frames and the \ac{mvs} network is called for each new keyframe. It would be possible to estimate depth only for every second keyframe to trade-off runtime and reconstruction quality.

Combining all the aforementioned measures decreases the inference time from 158 ms to 25.6 ms, a 83.2\% decrease, while maintaining reasonable reconstruction quality and without optimizations like Nvidia TensorRT. Additionally, the network could be used on every second keyframe, i.e. with ca. 2 Hz. This combination makes our \ac{mvsnet} $20\times$ real-time on an RTX 2080 super. The embedded Nvidia Jetson AGX Xavier has ca. 15\% of the computing power of the RTX 2080 super and thus our \ac{mvsnet} could potentially reach real-time on this device even without NVIDIA TensorRT. The classic SLAM components of TANDEM can be run in real-time on an embedded device \cite{DBLP:conf/ismar/SchopsEC14}, hence the proposed TANDEM could likely be implemented on an embedded device in real-time while maintaining reasonable reconstruction quality.

\end{document}